\documentclass{article} 
\usepackage{iclr2025_conference,times}

\usepackage{amsmath,amsfonts,bm}

\def\eqref#1{equation~\ref{#1}}

\def\1{\bm{1}}

\DeclareMathAlphabet{\mathsfit}{\encodingdefault}{\sfdefault}{m}{sl}
\SetMathAlphabet{\mathsfit}{bold}{\encodingdefault}{\sfdefault}{bx}{n}

\def\gD{{\mathcal{D}}}

\def\gP{{\mathcal{P}}}

\def\gT{{\mathcal{T}}}

\def\gX{{\mathcal{X}}}

\usepackage{multirow}
\usepackage{graphicx}
\usepackage{float}
\usepackage{overpic}
\usepackage{microtype}
\usepackage{inconsolata}
\usepackage{booktabs}
\usepackage{stfloats}
\usepackage{amsmath}
\usepackage{xcolor}

\usepackage{wrapfig}

\usepackage{hyperref}
\usepackage{url}

\usepackage{amssymb}

\let\emptyset\varnothing

\usepackage{algpseudocode}
\usepackage{amsthm}
\theoremstyle{definition}

\usepackage{epsfig,amsmath,amsfonts,epsfig,multirow,makecell,caption,soul,csquotes,color,wrapfig,mathtools,bm,spverbatim,booktabs,tcolorbox,diagbox}
\usepackage[e]{esvect}

\usepackage{caption}
\usepackage[first=0,last=9]{lcg}
\usepackage{colortbl}
\definecolor{Gray}{gray}{0.8}
\colorlet{Red}{red!10!white}
\colorlet{Blue}{blue!10!white}

\usepackage{hyperref}

\usepackage{scalerel}[2016/12/29]

\makeatletter
\providecommand{\leadsfrom}{%
  \mathrel{\mathpalette\reflect@squig\relax}%
}
\newcommand{\reflect@squig}[2]{%
  \reflectbox{$\m@th#1\leadsto$}%
}
\makeatother

\usepackage{amsmath,amsfonts,bm}

\def\eqref#1{equation~\ref{#1}}

\def\1{\bm{1}}

\DeclareMathAlphabet{\mathsfit}{\encodingdefault}{\sfdefault}{m}{sl}
\SetMathAlphabet{\mathsfit}{bold}{\encodingdefault}{\sfdefault}{bx}{n}

\def\gD{{\mathcal{D}}}

\def\gP{{\mathcal{P}}}

\def\gT{{\mathcal{T}}}

\def\gX{{\mathcal{X}}}

\usepackage{cutwin}
\usepackage{amsmath}
\usepackage{graphicx}
\usepackage{multirow}
\usepackage{algpseudocode}
\usepackage{algorithm}
\usepackage{amsfonts}       
\usepackage{nicefrac}       
\usepackage{microtype}      
\usepackage{xcolor}         
\usepackage[inline]{enumitem}
\usepackage{listings} 
\usepackage{xcolor} 
\usepackage{tcolorbox} 
\tcbuselibrary{breakable} 
\usepackage{lipsum}
\usepackage{listings}
\newtcolorbox{instructionbox}{
  colback=red!5!white,
  colframe=red!75!black,
  fonttitle=\bfseries,
  title=Instruction:,
  breakable
}
\newtcolorbox{answerbox}{
  colback=blue!5!white,
  colframe=blue!75!black,
  fonttitle=\bfseries,
  title=Answer:,
  breakable
}
\newtcolorbox{outputbox}{
  colback=green!5!white,
  colframe=green!75!black,
  fonttitle=\bfseries,
  title=Outputs:,
  breakable 
}
\newtcolorbox{mybox}{colback=white,
colframe=blue!75!black,
fonttitle=\bfseries,
title=Error Example,
  breakable}

  \newcommand{\Norm}[1]{\left\lVert#1\right\rVert}

\newcommand{\Abs}[1]{\left\lvert#1\right\rvert}

\usepackage{pifont}
\usepackage{xcolor}

\definecolor{DarkGreen}{RGB}{30,130,30}
\newcommand{\cmark}{\textcolor{DarkGreen}{\ding{51}}}
\newcommand{\xmark}{\textcolor{red}{\ding{55}}}%
\newcommand{\smark}{\textcolor{DarkGreen}{\ding{216}}}%
\newcommand{\hmark}{\textcolor{red}{\ding{218}}}%
\usepackage{booktabs}

\iclrfinalcopy

\title{Tool-Planner: Task Planning with Clusters across Multiple Tools}

\author{%
  Yanming Liu$^{1}$, Xinyue Peng$^{2}$, Jiannan Cao$^{3}$, Yuwei Zhang, Xuhong Zhang$^{1}$\footnotemark[1] ,\\ \bf
  Sheng Cheng$^{1}$, Xun Wang$^{1}$, Jianwei Yin$^{1}$, Tianyu Du$^{1}$\footnotemark[1]\ \\
  $^{1}$Zhejiang University, 
  $^{2}$Southeast University \\
  $^{3}$Massachusetts Institute of Technology \\
  \texttt{{\{oceann24, zhangxuhong, zjradty\}@zju.edu.cn, zjuyjw@cs.zju.edu.cn,}}\\
\texttt{{jiannan@mit.edu, xinyuepeng@seu.edu.cn, yuweizhang@tongji.edu.cn}} \\
}

\begin{document}

\maketitle

\renewcommand{\thefootnote}{\fnsymbol{footnote}}
\footnotetext[1]{Corresponding author.}
\renewcommand{\thefootnote}{\arabic{footnote}}

\begin{abstract}
  Large language models (LLMs) have demonstrated exceptional reasoning capabilities, enabling them to solve various complex problems. Recently, this ability has been applied to the paradigm of tool learning. Tool learning involves providing examples of tool usage and their corresponding functions, allowing LLMs to formulate plans and demonstrate the process of invoking and executing each tool. LLMs can address tasks that they cannot complete independently, thereby enhancing their potential across different tasks. However, this approach faces two key challenges. First, redundant error correction leads to unstable planning and long execution time. Additionally, designing a correct plan among multiple tools is also a challenge in tool learning. To address these issues, we propose Tool-Planner, a task-processing framework based on toolkits. Tool-Planner groups tools based on the API functions with the same function into a toolkit and allows LLMs to implement planning across the various toolkits. When a tool error occurs, the language model can reselect and adjust tools based on the toolkit. Experiments show that our approach demonstrates a high pass and win rate across different datasets and optimizes the planning scheme for tool learning in models such as GPT-4 and Claude 3, showcasing the potential of our method. Our code is public at \url{https://github.com/OceannTwT/Tool-Planner}.
\end{abstract}

\section{Introduction}

Large Language Models (LLMs)~\citep{GPT3, chowdhery2023palm, touvron2023llama, Zeng2023GLM, zhang2022opt} have demonstrated outstanding performance across multiple domains, leveraging parameterized knowledge to exhibit powerful reasoning and planning capabilities~\citep{bang2023multitask}. Tool learning~\citep{schick2024toolformer, yao2022ReACT, lu2024chameleon, qin2024toolllm} harnesses the planning prowess of LLMs by decomposing complex problems through understanding tools and having LLMs generate plans for tasks, thus leveraging external tools (APIs) to handle intermediate steps, achieving the goals of completing complex tasks. By utilizing tools, LLMs can significantly improve limitations in certain tasks, such as enhancing accuracy in mathematical reasoning problems~\citep{hao2024toolkengpt, lee2024math, gou2024tora}, answering up-to-date news queries~\citep{song2023restgpt, liu2024ra}, executing commands, or invoking other models~\citep{shen2024hugginggpt}. Consequently, tool learning has become one of the potential paradigms for solving complex real-world scenarios.

The existing study of tool learning focuses on two main crucial aspects: \textbf{(1) how to better utilize tools for task planning and execution}, and \textbf{(2) how to adjust planning based on the results of tool invocation.} During the process of invoking tools, chain-like invocations may encounter situations of failure~\citep{qin2024toolllm}. This necessitates timely adjustments of tools and task re-planning when encountering tool errors. DFSDT~\citep{qin2024toolllm} generates new plans by searching states when encountering API call errors, thus solving the original problem with tools according to new reasoning paths.  ToolChain*~\citep{zhuang2023toolchain} employs heuristic search algorithms to choose the direction most likely to yield answers during the planning search process. Tree-Planner~\citep{hu2023tree} presets multiple planning paths and merges nodes with the same preceding tool for deep search. These methods all involve adjustments and optimizations of search algorithms based on tree structures. However, when a tool is called and an error occurs, previous methods typically choose to discard that planning path directly, despite the potential existence of other tools offering similar or identical functionality to accomplish that task. Furthermore, the new planning methods proposed by LLMs are generally more complex, leading to a higher likelihood of tool errors on new planning paths. Methods like CRITIC~\citep{gou2024critic} and AnyTool~\citep{du2024anytool} integrate feedback mechanisms into tool learning and provide error messages to aid in task re-planning. Nonetheless, the integration of this information still faces issues of inefficiency while lacking effective exploration of the current planning path. Multiple re-planning cycles also result in inefficient task resolution. Therefore, in tool learning, the rational planning of tasks and tools becomes increasingly important.

\begin{table*}[!t]
    \centering
    \small
    \resizebox{0.95\linewidth}{!}{
    \begin{tabular}{lcccccc}
        \toprule
        \textbf{Feature} & \makecell{\textbf{Tool-Planner}\\(this work)} & \makecell{\textbf{ReAct}\\ \citep{yao2022ReACT}} & \makecell{\textbf{Reflexion}\\ \citep{shinn2024reflexion}} & \makecell{\textbf{AdaPlanner}\\ \citep{sun2024adaplanner}} & \makecell{\textbf{DFSDT}\\ \citep{qin2024toolllm}} & \makecell{\textbf{Toolchain*}\\ \citep{zhuang2023toolchain}} \\
        \cmidrule(lr){1-1} \cmidrule(lr){2-2} \cmidrule(lr){3-3} \cmidrule(lr){4-4} \cmidrule(lr){5-5} \cmidrule(lr){6-6} \cmidrule(lr){7-7}
        Tool Calling & \cmark & \cmark & \cmark & \cmark & \cmark & \cmark \\ 
        Tool Refinement & \cmark & \xmark & \cmark & \cmark & \cmark & \cmark \\ 
        Tool Expansion & \cmark & \xmark & \xmark & \xmark & \cmark & \cmark \\ 
        Explainable Planning & \cmark & \cmark & \xmark & \xmark & \cmark & \xmark \\ 
        Tool Integration & \cmark & \xmark & \xmark & \xmark & \xmark & \xmark \\ 
        Task Nodes Size & \smark & \hmark &\hmark & \hmark& \hmark & \hmark \\ 
        \bottomrule
    \end{tabular}
    }
    \caption{
    \small{A comparison of our \textbf{Tool-Planner} to notable planning and tool interaction frameworks. Our method shows significant advantages in real-world tool integration, multi-tool support, and efficient solution sizes.}
    }
    \label{tab:method_comparison}
\end{table*}

To address these challenges, we propose Tool-Planner, an efficient framework for task planning and tool invocation in tool learning with LLMs. Tool-Planner conceptualizes the problem-solving process as a decision tree. Unlike the previous methods, Tool-Planner views each node as a set of tools rather than a single tool, when a tool invocation error occurs, it prioritizes solutions within the same toolkit. Tool-Planner effectively tackles the inefficiency issue of previous methods in utilizing task planning solutions and significantly improves the pass rate of tool learning in task resolution. We construct sets of similar tools utilizing the SimCSE~\citep{gao2021simcse} to evaluate the distance of different tools for tool clustering based on tool APIs document and description. We conduct extensive experiments on two different datasets: ToolBench~\citep{qin2024toolllm}, which uses APIs selected from RapidAPI Hub \citep{rapidapi}, and APIBench~\citep{patil2023gorilla}, which fetches APIs from various open-source models.  Compared to various prior tool-learning search methods, Tool-Planner achieves a +8.8\% increase in pass rate and +9.1\% increase in win rate on ToolBench, as well as a +6.6\% increase in pass rate and +14.5\% increase in win rate on APIBench when tested with GPT-4. It also demonstrates outstanding performance in terms of re-planning frequency and computational speed. Extensive experimental results highlight the advancements of Tool-Planner.

\textbf{Our Contributions.} Our main contributions are summarized as follows.
\begin{itemize}[leftmargin=*]
\item We propose a novel framework, Tool-Planner, which integrates external tools with LLMs. This framework enables task planning and tool invocation based on toolkits, addressing the inefficiencies in planning found in previous approaches.

\item Tool-Planner categorizes tools into toolkits with similar or identical functionalities by clustering tool embeddings generated by SimCSE. The setting of toolkits allows thorough exploration of tools along a planning path, ensuring that each node is fully utilized and maximizing the information from that path.

\item Extensive experimental results demonstrate the effectiveness of the Tool-Planner framework, highlighting the importance of thorough tool planning exploration on a single path compared to DFSDT's multiple attempts at different planning paths, and providing an ideal paradigm for future tool learning solution space search methods.
\end{itemize}

\section{Preliminaries}

\textbf{LLMs Reasoning with Tools.} Given a task $x$ input in natural language and a pre-trained model $\rho_{\theta}(x)$, the naive generation output of an LLM is $y \sim \rho_{\theta}(x)$, corresponding to the answer predicted by the model. In the context of tool learning with LLMs \citep{lu2024chameleon, qin2024toolllm}, given the API documentation (or demos) for tools $\mathcal{D}=\{d_i\}_{i=1}^{N}$ and their API descriptions $\mathcal{M}=\{m_i\}_{i=1}^{N}$, we first generate a multi-step decision-making plan for the tools $\mathcal{P}_t = \{p_1, p_2, …, p_N\} \sim \rho_{\theta}(x, \mathcal{D}, \mathcal{M})$, where $N$ is the number of tools in the plan. Thus, for each intermediate reasoning step $x_i$, it can be generated through a series of corresponding tool calls. Setting the result of the API function call for each tool as $c_i = F_i(p_i)$, we have $x_i = \rho_{\theta}(x_i|\{x_h\}_{h=1}^{i}, \{c_{{\mathcal{T}}_h}\}_{h=1}^{i-1}, \mathcal{D}, \mathcal{M}, x)$. Consequently, the final output result or behavior for the task can be expressed as $y = \rho_{\theta}(y|\{x_h\}_{h=1}^{K}, \{c_{{\mathcal{T}}_h}\}_{h=1}^{K}, \mathcal{D}, \mathcal{M}, x)$. Such a process of tool invocation significantly enhances the adaptability of LLMs to various tasks.

\textbf{Tree Search on Planning Space.} 
In the chain-like calls of tool learning \citep{yao2022ReACT}, the use of tools is linear. When encountering the hallucination problem of LLMs leading to repeated API calls and parameter errors, or when the tool itself is unavailable, the exploration of the solution space $\mathcal{S}$ is insufficient. This requires us to adjust the planned path $\mathcal{P}_t$ accordingly. The exploration of the planned path can be seen as a behavior tree, formalized as $G(c) = (V, \mathcal{E})$. Whenever re-planning is needed, all previous plan paths $\mathcal{P}$ and the original plan are used to generate a new plan tree chain $\mathcal{P}_{t'} = \{p_h'\}_{h=1}^K  \sim \rho_{\theta}(x, \mathcal{D}, \mathcal{M}, \mathcal{P}, E)$ based on the returned error information $E = \sum \{F_{e}(p_{e})\}$, and priority is given to finding the solution states that can be reached after correcting the current state. However, such multiple iterative calls and exploration of the solution space may repeatedly encounter hallucinations and tool issues, leading to excessively high costs in exploring solutions.

\section{Methodology}
\label{headings}

When LLMs utilize various tools to generate answers, inefficiencies often emerge from a suboptimal selection process among multiple tools and frequent alterations in problem-solving strategies. To address this issue, we propose the Tool-Planner framework, aiming at enhancing the efficiency of tool calls while maintaining relative consistency in planning. Our framework includes clustering and categorizing tools, as well as planning processes for tool invocation paths.

\subsection{Tool Clustering}
\label{sec:toolcluster}

In the realm of tool learning, comprehending the functionalities of application programming interfaces (APIs) is essential. However, the functionalities of most APIs are often inadequately categorized. For instance, platforms like RapidAPI group APIs with different functions into the same general category without categorizing them specifically for particular tasks. Consequently, problem-solving may require searching through multiple paths and invoking similar functionalities from different categories, leading to increased complexity and higher error rates in implementation.

\begin{wrapfigure}{r}{6.5cm} 
\includegraphics[width=6cm]{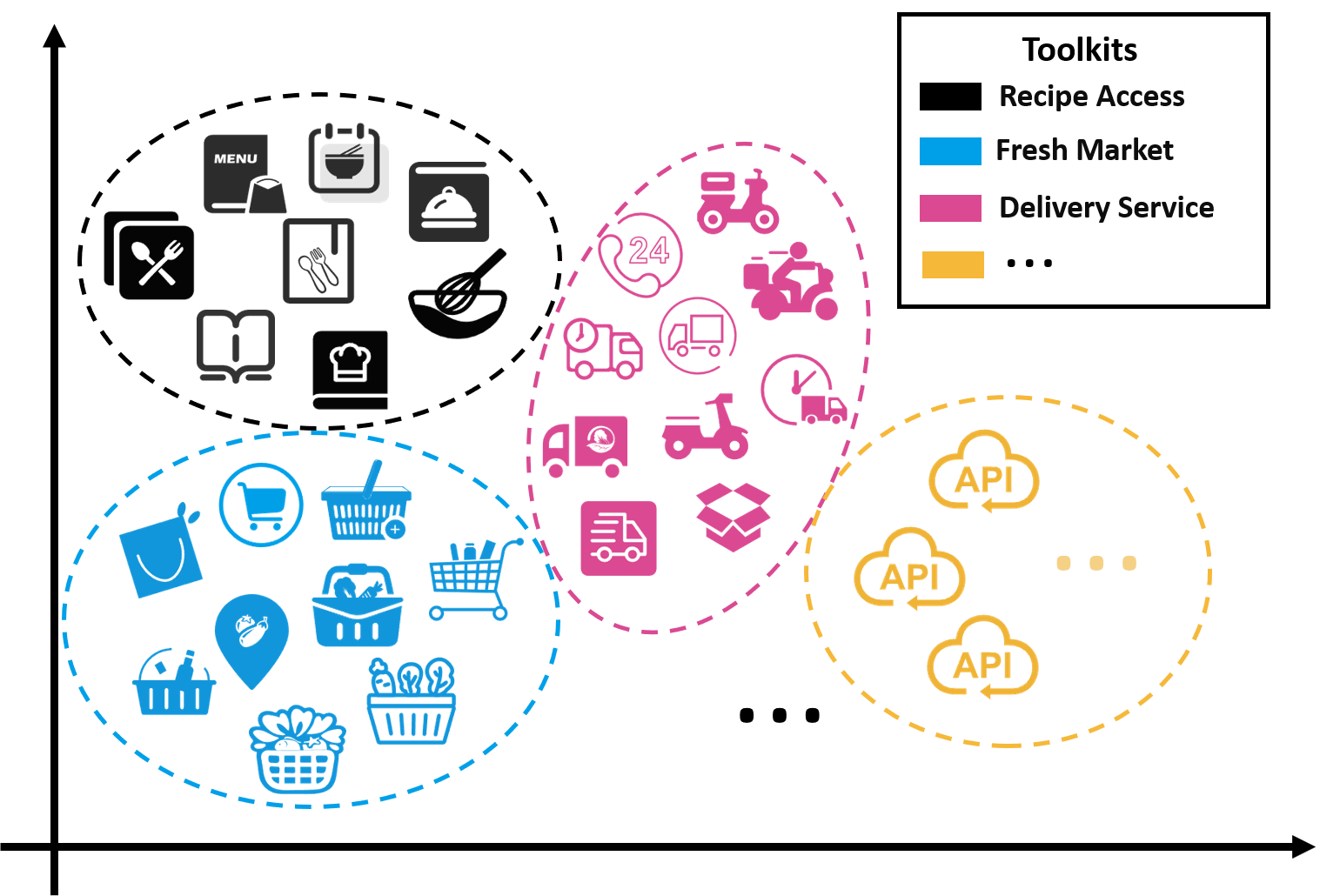}
\caption{The process of tool clustering.}
\label{fig:tibet}
\end{wrapfigure}
To tackle this issue, we need to categorize a large number of tools into multiple classes $\mathcal{T}_i$ with similar or identical functionalities, providing functionality explanations for each class, as shown in Figure~\ref{fig:tibet}. Through annotated categorization, each class is assigned to a specific functionality $f(i)$, we can accurately differentiate APIs with different functionalities, enabling the selection of alternative APIs from the same category if the chosen API fails.

However, as the number of APIs may continue to increase, annotating each API is costly, and accurately identifying categories from multiple classes poses a challenge. 
It has been proven effective to use LLMs to guide text and assist in clustering different texts \citep{viswanathan2024large, tipirneni2024context, zhang2023clusterllm}. Therefore, we develop an automated classification method. We extracted tool documentation $\mathcal{D}$ and descriptions $\mathcal{M}$ of candidate APIs and provided them to LLMs to generate brief explanations $H = \{h_i\}_{i=1}^N$ of their functionalities. Upon obtaining API explanations, we utilize the SimCSE \citep{gao2021simcse} model to compute text embeddings for these explanations, which can be seen as tool embeddings $e_i$ for the API. To understand which tools might have similar functionalities, we need to classify tool embeddings.

After obtaining tool embeddings, we employ the $k$-means++ \citep{arthur2007k} algorithm to find a $k$-partition of these tool embeddings and generate tool clusters in the solution space. This could formulated as an optimization problem:
\begin{equation}
\arg\min_{\mathcal{T}}\sum_{i=1}^k\sum_{e\in \mathcal{T}_i}\Norm{e-\frac{1}{\Abs{\mathcal{T}_i}}\sum_{x\in \mathcal{T}_i}x}_2^2.  
\end{equation}

The number of clusters is adjusted through the value of $k$. Each API is assigned to a tool cluster with similar or identical functionalities, even if they are not in the same tool within RapidAPI. Similar functionalities enable rapid tool adjustment when addressing issues. We refer to a cluster as a toolkit, meaning that one toolkit can perform a specific task.

\begin{figure*}[tp]
    \centering
\includegraphics[width=1\linewidth]{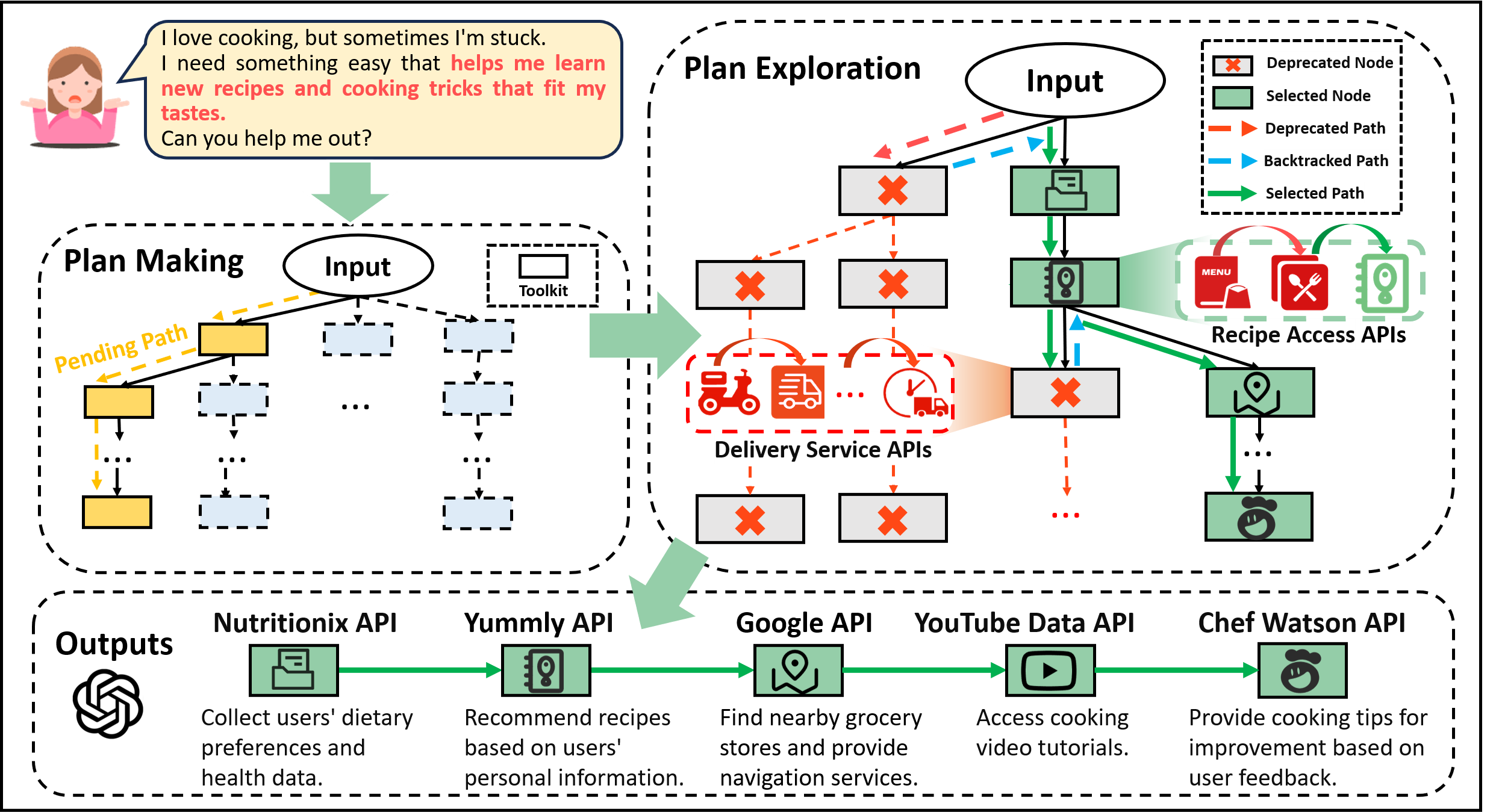}
    \caption{Overview of Tool-Planner. The user inputs a problem, the LLM devises a plan with API functionalities as nodes. The plan is then explored step by step. If an error occurs, other APIs within the same toolkit are tried. If all APIs in the toolkit fail, the process moves to other nodes until the final plan is determined and output.
    }
    \label{fig:overview}
\end{figure*}

\subsection{Task Planning}

Given task input $x$, LLMs first provide a plan $\mathcal{P}$ for the task. This approach differs from previous methods that directly presented API documentation to LLMs, leading to excessively lengthy contexts. In contrast, Tool-Planner first utilizes in context learning \citep{GPT3} to generate brief explanations for all API functionalities within the toolkit, and then uses these explanations to generate the functionality descriptions of toolkit $\mathcal{T}_h$. Consequently, each toolkit is associated with a unique functionality description, denoted as $m_{\mathcal{T}_h} = f(\mathcal{T}_h)$. When planning tasks, we provide these toolkit functionality descriptions as context to LLM and let it design a plan  $\mathcal{P}_{\mathcal{T}} = \{P_h\}_{h=1}^{N} \sim \rho_{\theta}(x, \{m_{\mathcal{T}_h}\}_{h=1}^k)$ using different toolkits to solve the task. With this prompt, the LLM can generate a chained toolkit-based plan based on the functionality descriptions of the toolkits.

In solving problems for specific states, the model will choose any API within the toolkit for invocation, thus completing the problem-solving for that state and passing the output result to the next state. While the chosen API or tool is $t$, the return of each intermediate toolkit could be stated as $c_{{\mathcal{T}}_i} = F_t(t)$, we have the intermediate state for instruction:
\begin{equation}
x_i = \rho_{\theta}(x_i|\{x_h\}_{h=1}^{i-1}, \{c_{{\mathcal{T}}_h}\}_{h=1}^{i-1}, \{m_{{\mathcal{T}}_h}\}_{h=1}^i, \mathcal{D}_\mathcal{T}, x).
\end{equation}

In this way, when APIs within each toolkit can function properly, this plan can choose any API for each state to complete this process, and provides multiple choices of APIs to complete a state.

\subsection{Planning Exploration on Solution Space}

When LLMs experience hallucinations resulting in problematic parameter information or tool unavailability, we need to replan to complete the original task. After categorizing various tools, since each node in our search plan represents a toolkit containing multiple available APIs, we can adjust the tools through the following adaptation. The behavior tree in the toolkit plan could be formalized as $G(\mathcal{T}) = (V_\mathcal{T}, \mathcal{E}_\mathcal{T})$.

\textbf{Task Planning Within Same Toolkits. }When the current API $t$ within a node $V_\mathcal{T}$ becomes unusable for any reason, we prioritize selecting another available API $t'$ within the same toolkit. By referring to the API documentation $d_{t'}$, we can generate the call parameters $param$ for the new API, and fetch the calling result $c'_{{\mathcal{T}}_h} = F_t(param)$. This allows us to complete the current state by selecting an alternative tool within the same toolkit, without altering the original task plan. In other words, keep \( \mathcal{E}_\mathcal{T} \) unchanged. The results generated by the new API can then be used as input for the next state in the task plan, maintaining the relative stability of the plan. The intermediate state for instruction is:
\begin{equation}
x_i = \rho_{\theta}(x_i|\{x_h\}_{h=1}^{i-1},\{c_{{\mathcal{T}}_h}\}_{h=1}^{i-1}, \{c'_{{\mathcal{T}}_{i}}\}, \{m_{{\mathcal{T}}_h}\}_{h=1}^i, \mathcal{D}_\mathcal{T}, x).
\end{equation}

\textbf{Task Replanning Across Toolkits. }Given the limited number of identical or similar APIs within a toolkit, if all APIs fail to process the current state $v$, we provide all error information $E = \sum_{t \in \mathcal{T}_i} \{F_{t}(t_{e})\}$ and the original task plan $\mathcal{P}_\mathcal{T}$ to the LLMs, instructing them to generate a new task plan $\mathcal{P}'_\mathcal{T}$ on subgraph $G(\mathcal{T}) \leftarrow G(\mathcal{T}) \setminus {v}$.  The new task plan aims to retain as many results from the previous state as possible. This process is similar to DFSDT \citep{qin2024toolllm}, but while DFSDT searches at the API level, we search at the toolkit level. Once a new plan is generated, we select APIs within the new toolkit and attempt to complete the new state $v'$ according to the new toolkit's APIs. In this way, we can switch to a new plan when the original plan completely fails to accomplish the task, while maintaining relative small times to generate a new plan.
\begin{equation}
\mathcal{P}'_\mathcal{T} = \{P_h'\}_{h=1}^{K} \sim \rho_{\theta}(x, \mathcal{D}, \{m_{{\mathcal{T}}_h}\}_{h=1}^K, \mathcal{P}_\mathcal{T}, E), \text{on} \  G(\mathcal{T}) \leftarrow G(\mathcal{T}) \setminus {v}.
\end{equation}
By setting these two possibilities, we can greatly optimize the task planning process, thereby improving the pass rate and effectiveness of task completion. We demonstrate more details on Appendix~\ref{title:C}.

\section{Experiment}

To evaluate the capabilities of our framework, we conducted a series of experiments and assessments, demonstrating the superiority of our approach from various aspects. 

\subsection{Experiment Setup}

\textbf{Dataset.} We utilize the ToolBench \citep{qin2024toolllm} and APIBench\citep{patil2023gorilla} as our experimental dataset. ToolBench comprises 16,464 APIs, which are categorized into different tools and categories. 
In ToolBench, there are three different datasets for prompt generation, namely G1, G2, and G3, which represent single-tool instructions, intra-category multi-tool instructions, and intra-collection multi-tool instructions, respectively. In APIBench, datasets are constructed by selecting corresponding 1,645 APIs and their descriptions from three platforms as tools, and a series of questions are used to evaluate their performance. More details are described in Appendix~\ref{title:B}. We use the API interfaces selected by ToolBench and APIBench along with their corresponding documentation and descriptions to extract and generate functional explanations of the APIs. Subsequently, we generate tool embeddings based on their functionalities $\{m_{\mathcal{T}_h}\}_{h=1}^k$. 

\textbf{Evaluation Metrics.} For the ToolBench dataset, we adopt two metrics from ToolEval \citep{qin2024toolllm} to evaluate our framework, covering different aspects of the task. The first metric is \textbf{Pass Rate}, calculated based on the proportion of tasks successfully completed. The second metric is \textbf{Win Rate}, where we compare the solution generated by our method with the plan generated by GPT-3.5+ReACT and make LLMs judge which solution is better. When our framework performs better, we mark it as a win. If our framework is the same or inferior to the GPT3.5+ReACT solution, we mark it as a tie or loss. The win rate reflects the quality of our generated solutions and their ability to solve problems. We also evaluate APIBench in \textbf{Halluation Rate}. Since most of the call processes in APIBench involve only a few tools, we can make comparison on the hallucination situation of the reasoning process. 

\textbf{Baselines. } We compare Tool-Planner with the following baseline methods. (1) \textbf{ReACT} \citep{yao2022ReACT} operates by having an LLM execute an action based on the previous state, and then reason based on the result of that action, repeating this process iteratively. This can be considered a linear tool usage process. (2) \textbf{Reflexion} \citep{shinn2024reflexion} introduces a feedback mechanism during decision-making. In tool-learning scenarios, when an error occurs with an intermediate tool, Reflexion searches for the next node based on previous nodes using the error information. (3) \textbf{AdaPlanner} \citep{sun2024adaplanner} is similar to Reflexion, which explicitly corrects an error at a specific point in the path and adjusts by selecting the correct tool. (4) \textbf{DFSDT} \citep{qin2024toolllm} employs a deep search mechanism. Each time an error node is encountered, it provides the model with all previous error paths and information, allowing the model to re-select and re-plan the path, thereby maximizing the expansion of possible solutions.

\textbf{Model.} We utilize the following foundational models for model planning generation and tool learning problem-solving: GPT-3.5 \citep{ouyang2022training} (\textit{gpt-3.5-turbo-0125}), GPT-4 \citep{achiam2023gpt} (\textit{gpt-4-turbo-2024-04-09}), and Claude-3 (\textit{claude-3-sonnet}). We use SimCSE \citep{gao2021simcse}, an effective method for sentence representation learning that leverages contrastive learning to calculate the tool embeddings. We set $k$ as 1800 in ToolBench and $k$ as 65 in APIBench for experiments.

\begin{table*}[]
    \centering
    \caption{The results of different models and baselines of Pass Rate (\%), Win Rate (\%), and Halluation Rate (\%). We evaluate various methods on three sub-dataset with their APIs derived from APIBench\citep{patil2023gorilla}.}
    \label{tab:main_result}
    \resizebox{1.0\textwidth}{!}{
    \begin{tabular}{llcccccccccccc}
    \toprule
    \multirow{2}{*}{\textbf{Model}}                       & \multirow{2}{*}{\textbf{Method}} & \multicolumn{3}{c}{\textbf{TorchHub}} & \multicolumn{3}{c}{\textbf{HuggingFace}} & \multicolumn{3}{c}{\textbf{TensorFlow}} & \multicolumn{3}{c}{\textbf{Average}} \\ 
    \cmidrule(lr){3-5} \cmidrule(lr){6-8} \cmidrule(lr){9-11} \cmidrule(lr){12-14}
                                                 &                         & Pass.($\uparrow$)         & Win.($\uparrow$)          & Hallu.($\downarrow$)         & Pass.($\uparrow$)         & Win.($\uparrow$)         & Hallu.($\downarrow$)         & Pass.($\uparrow$)          & Win.($\uparrow$)         & Hallu.($\downarrow$)         & Pass.($\uparrow$)          & Win.($\uparrow$)          & Hallu.($\downarrow$)        \\ 
    \midrule
    \multirow{5}{*}{GPT-3.5}                    & ReACT                   & 62.6            &      -        & 23.2          & 35.6            &      -       & 31.6          & 53.7            &  -           & 13.6          & 50.6         & -            & 22.8          \\
    \multicolumn{1}{c}{}                         & Reflexion              & 70.2           & 59.0         & 21.7          & 46.0            & 55.2          & 28.0          & 55.0           & 56.5            & 12.7          & 57.1         & 56.9        & 20.8          \\
    \multicolumn{1}{c}{}                         & AdaPlanner              & 69.5          & 62.1            & 23.7          & 48.7             & 53.4           & 32.7          & 57.2            & 53.6           & 14.2          & 58.5         & 56.4        & 23.5          \\
    \multicolumn{1}{c}{}                         & DFSDT                   & 74.5           & 72.0        & 9.6          & 57.2          & 60.7         & 18.2          & 70.9           & 69.6           & \textbf{11.2}          & 67.5         & 67.4        & 13.0          \\
    \multicolumn{1}{c}{}                         & \cellcolor{Red}Tool-Planner       & \cellcolor{Red}\textbf{78.2}             & \cellcolor{Red}\textbf{77.6}            & \cellcolor{Red}\textbf{8.2}          & \cellcolor{Red}\textbf{66.5}         & \cellcolor{Red}\textbf{75.4}           & \cellcolor{Red}\textbf{13.5}          & \cellcolor{Red}\textbf{72.5}           & \cellcolor{Red}\textbf{72.4}           & \cellcolor{Red}12.1          & \cellcolor{Red}\textbf{72.7}           & \cellcolor{Red}\textbf{75.1}           & \cellcolor{Red}\textbf{11.3}           \\ 
    \midrule
    \multirow{5}{*}{GPT-4}                       & ReACT                   & 68.6            &      58.0        & 19.2          & 43.8            &      59.2       & 23.4          & 61.9            &  61.1          & 15.8          & 58.1         & 59.4            & 19.4          \\
                                                 & Reflexion              & 74.4           & 68.4         & 16.5          & 51.9            & 58.2          & 16.8          & 60.2           & 55.5            & 12.1          & 62.2         & 60.7        & 15.1          \\
                                                 & AdaPlanner              & 75.3          & 78.5            & 15.9          & 50.2             & 56.8           & 15.3          & 60.1            & 53.9           & 12.7          & 61.8         & 63.1       & 14.6          \\
                                                 & DFSDT                   & 80.3           & 78.1        & 11.2          & 58.4          & 62.0         & 12.3          & 72.1           & 69.0           & 10.9          & 70.3         & 67.7        & 11.5          \\
                                                 &\cellcolor{Red}Tool-Planner                   & \cellcolor{Red}\textbf{83.2}             & \cellcolor{Red}\textbf{87.0}            & \cellcolor{Red}\textbf{7.1}          & \cellcolor{Red}\textbf{70.3}         & \cellcolor{Red}\textbf{77.0}           & \cellcolor{Red}\textbf{10.7}          & \cellcolor{Red}\textbf{77.2}           & \cellcolor{Red}\textbf{82.6}           & \cellcolor{Red}\textbf{10.5}          & \cellcolor{Red}\textbf{76.9}           & \cellcolor{Red}\textbf{82.2}           & \cellcolor{Red}\textbf{9.4}           \\ 
    \midrule
    \multirow{5}{*}{Claude-3}                    & ReACT                   & 69.5            &      56.5        & 21.6          & 42.1            &      56.9       & 30.1          & 58.9           &  46.8          & 13.2          & 56.8         & 53.4            & 21.6          \\
                                                 & Reflexion              & 76.8           & 68.5         & 20.1          & 48.8            & 67.1          & 20.4          & 60.3           & 59.1            & 12.4          & 62.0         & 64.9        & 17.6          \\
                                                 & AdaPlanner              & 75.2          & 69.8            & 18.4          & 49.5             & 65.6           & 16.5          & 62.7            & 61.3           & 13.8          & 62.5         & 65.6        & 16.2         \\
                                                 & DFSDT                   & 80.2           & 72.6        & 10.0          & 55.8          & 69.2         & 14.9          & \textbf{73.4}          & 65.0           & 11.5          & 69.8         & 68.9        & 12.1          \\
                                                 &\cellcolor{Red}Tool-Planner                   & \cellcolor{Red}\textbf{81.5}             & \cellcolor{Red}\textbf{83.5}            & \cellcolor{Red}\textbf{8.6}          & \cellcolor{Red}\textbf{68.2}         & \cellcolor{Red}\textbf{75.9}           & \cellcolor{Red}\textbf{12.4}          & \cellcolor{Red}72.1           & \cellcolor{Red}\textbf{74.5}           & \cellcolor{Red}\textbf{11.1}          & \cellcolor{Red}\textbf{73.9}           & \cellcolor{Red}\textbf{78.0}           & \cellcolor{Red}\textbf{10.7}           \\ 
    \midrule
    \end{tabular}
    }
\end{table*}

\subsection{Main Experiment}

\textbf{For APIBench.} Tool-Planner outperforms other methods across all three API platforms with significant improvements in Pass Rate, Win Rate, and Hallucination Rate. Compared to DFSDT, Tool-Planner increases the pass rate by \textbf{+6.6\%} and the win rate by \textbf{+14.5\%}, while reducing the hallucination rate by \textbf{-2.1\%} on GPT-4. This demonstrates that Tool-Planner's clustering-based tool planning approach better captures the nuances of API and model selection and task execution, leading to more robust and reliable performance.

\textbf{Model Invocation and Hallucination.} Tool-Planner significantly reduces the occurrence of hallucinations. This is due to the targeted optimization of the current model's performance during multiple attempts and plan revisions. For plans that fail to invoke the API successfully, the model discards those plans and selects another API that can be called. The plans generated through this approach effectively invoke the corresponding APIs of the model, successfully achieving the intended goals. Models that successfully invoke APIs also gain better interpretation within toolkit.

\textbf{For ToolBench.} Tool-Planner achieves state-of-the-art performance on five out of six datasets and demonstrates competitive performance on G1-Cat. Table~\ref{tab:main_result} shows the comparison results between Tool-Planner and other baselines. Compared to the DFSDT, which has the best average performance, our method improves the pass rate by \textbf{+8.8\%} and the win rate by \textbf{+9.1\%}. This indicates that by clustering tools and planning on the clustered toolkits, task planning capabilities can be more effectively enhanced.

\textbf{Multi-tool instructions tasks.} For tasks that require the cooperation of multiple tools (such as G2 and G3), our method shows remarkable performance improvements. For multi-tool instructions tasks, the pass rate increases by \textbf{+8.9\%} and the win rate by \textbf{+10.6\%} on GPT4, which is more notable than the improvement in single-tool scenarios. This demonstrates that our method is better able to adapt to and leverage the advantages of different toolkits in multi-tool scenarios. The setup of the toolkits ensures that tool coordination in multi-step tasks focuses more on the execution of each step and can find suitable solutions within similar tools when encountering errors, ensuring the relative stability of the plan.

\begin{table*}[]
\centering
\caption{The results of different models and baselines of Pass Rate (\%) and Win Rate (\%). We evaluate the results on six different sub-dataset derived from Toolbench with various methods.
}
\label{tab:apibench_main_result}
\
\resizebox{1.0\textwidth}{!}{
\begin{tabular}{llllllllllllllll}
\toprule
\multirow{2}{*}{\textbf{Model}}                       & \multirow{2}{*}{\textbf{Method}} & \multicolumn{2}{c}{\textbf{G1-Inst.}} & \multicolumn{2}{c}{\textbf{G1-Tool.}} & \multicolumn{2}{c}{\textbf{G1-Cat.}} & \multicolumn{2}{c}{\textbf{G2-Inst.}} & \multicolumn{2}{c}{\textbf{G2-Cat.}} & \multicolumn{2}{c}{\textbf{G3-Inst.}} & \multicolumn{2}{c}{\textbf{Avg.}} \\ \cmidrule(lr){3-4} \cmidrule(lr){5-6} \cmidrule(lr){7-8} \cmidrule(lr){9-10} \cmidrule(lr){11-12} \cmidrule(lr){13-14} \cmidrule(lr){15-16} 
                                             &                         & Pass.         & Win.         & Pass.         & Win.        & Pass.         & Win.         & Pass.         & Win.         & Pass.         & Win.         & Pass.         & Win.         & Pass.       & Win.       \\ \midrule
\multirow{5}{*}{GPT-3.5} & ReACT                   &   39.5            &      -        &   40.5            &      -       &    40.0           &      -        &   33.5            &   -           &  30.5             &      -        &    21.0           &    -          &    34.2         &    -        \\
\multicolumn{1}{c}{}                         & Reflexion              &    45.5           &   55.8           &   55.0            &   58.8          &    53.5           &  55.5            &    58.0           &    56.3          &  66.0             &  57.3            &     55.0           &    61.3          &     55.5        &   57.5         \\
\multicolumn{1}{c}{}                         & AdaPlanner              &     46.0          &  56.3            &  53.5             &  60.5           &     55.0          &   57.3           &   60.5            &  65.8            &   67.5            &  55.8            &   59.0            &  59.5            &    56.9         &     59.2       \\
\multicolumn{1}{c}{}                         & DFSDT                   &    48.5           &      58.3        &     62.5          &    59.8         &     58.0          &    56.8          &         71.5      &     70.5         &    70.5           &     59.3         &     61.0          &        \textbf{68.0}      &      62.0       &     62.1       \\ 
\multicolumn{1}{c}{}                         & \cellcolor{Red}Tool-Planner       &    \cellcolor{Red}\textbf{58.5}             &  \cellcolor{Red}\textbf{63.3 }            &    \cellcolor{Red}\textbf{71.0 }         &   \cellcolor{Red}\textbf{63.5 }           &    \cellcolor{Red}\textbf{66.0}        &    \cellcolor{Red}\textbf{61.3}          &   \cellcolor{Red}\textbf{75.5}           &    \cellcolor{Red}\textbf{72.3}           &   \cellcolor{Red}\textbf{78.0}           &   \cellcolor{Red}\textbf{63.5}           &    \cellcolor{Red}\textbf{66.0}          &         \cellcolor{Red}67.5      &    \cellcolor{Red}\textbf{69.2}          &     \cellcolor{Red}\textbf{65.2}       \\ \midrule
\multirow{5}{*}{GPT-4}                       & ReACT                   &  50.5             &   59.8           &    46.5           &   59.3          &    51.5           &   63.0           &        64.5       &    62.8          &   67.5            &   58.5           &   42.0            &       73.5       &    53.8         &    62.9        \\
                                             & Reflexion    & 49.5         &    60.8           &  56.5            &  65.3             &  68.0           &    68.5           &   74.0           &   71.8            &    67.0          &   55.3            &  55.0            &    76.5           &    61.7          &     66.4                    \\
                                             & AdaPlanner             &     52.5          &   62.5           &  60.5             &  64.8           &   68.5            &  \textbf{73.3}               &   75.5           &    73.5           & 68.0             & 52.8              &   70.0           &   79.5          &    65.9 &    67.7    \\
                                             & DFSDT                   &   57.0            &   65.8           &  72.0            &   69.3          &   64.5            &   65.3           &  77.5             & 72.0             &    69.5           &  56.8            &   71.0            &   81.5           &   68.5          &    68.4    \\ 
                                             & \cellcolor{Red}Tool-Planner                   &    \cellcolor{Red}\textbf{66.0}           &   \cellcolor{Red}\textbf{75.5}           &     \cellcolor{Red}\textbf{78.5}          &  \cellcolor{Red}\textbf{75.8}          &   \cellcolor{Red}\textbf{75.0}            &   \cellcolor{Red}71.8           &   \cellcolor{Red}\textbf{83.5}            &    \cellcolor{Red}\textbf{79.8}          &  \cellcolor{Red}\textbf{77.5}             &    \cellcolor{Red}\textbf{70.3}          &     \cellcolor{Red}\textbf{83.0}          & \cellcolor{Red}\cellcolor{Red}\textbf{92.0}             &  \cellcolor{Red}\textbf{77.3}           &   \cellcolor{Red}\textbf{77.5}          \\ \midrule
\multirow{5}{*}{Claude-3}                    & ReACT                   &  48.5             &   58.8           &  47.5             &    57.3         &    50.0           &  62.5            &   63.0            &  61.5            &    62.5           &   54.8           &  38.0             &      68.5        &  51.6           &     60.6       \\
                                             & Reflexion              &  51.5             &    62.3          &    58.5           &    65.8         &  55.5             & 64.3             &    73.0           &   72.8           &    67.0           &    54.3          &     61.0          &   73.0           &   61.1          &   65.4         \\
                                             & AdaPlanner              &     52.0          &   63.5           &  62.5             &  64.3           &   57.0            &  63.8           &  71.5             &  73.3            &   66.5            & 53.3             &   62.0            &        75.5      &   61.9          &  65.6          \\
                                             & DFSDT                   &   55.5            &  62.3            &   72.5            &  68.5           &    \textbf{61.5}           &  67.5            & 74.5              &  70.3            &    68.0           &       56.8       &     69.0          &     79.0         &   66.8          &    67.4        \\ 
                                             & \cellcolor{Red}Tool-Planner                   &    \cellcolor{Red}\textbf{64.0}           &  \cellcolor{Red}\textbf{73.8}            &    \cellcolor{Red}\textbf{77.0}           &    \cellcolor{Red}\textbf{76.3}         &  \cellcolor{Red}59.5             &   \cellcolor{Red}\textbf{73.8}           &         \cellcolor{Red}\textbf{79.5}      &   \cellcolor{Red}\textbf{79.3}           & \cellcolor{Red}\textbf{76.5}              &  \cellcolor{Red}\textbf{68.3}           &    \cellcolor{Red}\textbf{78.0}           &    \cellcolor{Red}\textbf{87.5}          &   \cellcolor{Red}\textbf{72.4}          &     \cellcolor{Red}\textbf{76.5}       \\ \midrule
\end{tabular}
}
\end{table*}

\subsection{Analysis}

\textbf{Ablation study on tool clustering.} In Section~\ref{sec:toolcluster}, we introduced the method of tool clustering, which is applied during the phase of generating plans by the model. At each planning step, we obtain a variety of APIs with similar functionalities through clustering as alternatives. To validate the effectiveness of tool clustering in the Tool-Planner, we combine the baselines ReACT and AdaPlanner with the toolkit obtained through clustering, and conducted experiments on GPT-4. Subsequently, we calculated their pass rate and win rate, with detailed results shown in Table~\ref{tab:toolkit}.

\begin{wraptable}[16]{r}{0.6\linewidth}
	\centering
 \caption{Results of Pass Rate (\%) and Win Rate (\%) improvement with tool clustering algorithm integration using GPT-4 across various baselines.}
 \label{tab:toolkit}
 \centering
\resizebox{\linewidth}{!}{
    \begin{tabular}{@{}cllllllll@{}}
    \toprule
    \multirow{2}{*}{\textbf{Method}} & \multicolumn{2}{c}{\textbf{G1-Tool.}} & \multicolumn{2}{c}{\textbf{G2-Inst.}} & \multicolumn{2}{c}{\textbf{G2-Cat.}} & \multicolumn{2}{c}{\textbf{G3-Inst.}} \\ \cmidrule(l){2-3} \cmidrule(l){4-5} \cmidrule(l){6-7} \cmidrule(l){8-9}
     & Pass. & Win. & Pass. & Win. & Pass. & Win. & Pass. & Win. \\ 
     \midrule
    \multicolumn{9}{c}{\textbf{\textit{w/o Toolkit Integration}}} \\ 
    \midrule
    ReACT & 46.5 & 59.3 & 64.5 & 62.8 & 67.5 & 58.5 & 42.0 & 73.5 \\
    AdaPlanner & 60.5 & 64.8 & 75.5 & 73.5 & 68.0 & 52.8 & 70.0 & 79.5 \\
    DFSDT & 72.0 & 69.3 & 77.5 & 72.0 & 69.5 & 56.8 & 71.0 & 81.5 \\ 
    \midrule
    \multicolumn{9}{c}{\textbf{\textit{with Toolkit Integration}}} \\ 
    \midrule
    ReACT & 55.5 & 62.3 & 72.5 & 69.0 & 68.0 & 61.3 & 59.0 & 77.8 \\
    AdaPlanner & 72.5 & 71.3 & 78.5 & 76.0 & 70.0 & 57.5 & 73.0 & 83.0 \\
    Tool-Planner & \textbf{78.5} & \textbf{75.8} & \textbf{83.5} & \textbf{79.8} & \textbf{77.5} & \textbf{70.3} & \textbf{83.0} & \textbf{92.0} \\ 
    \bottomrule
    \end{tabular}}
\end{wraptable}

The experimental results indicate that the performance significantly improved after applying our tool clustering algorithm, even under a single-chain approach. \textbf{Tool clustering helps provide alternative solutions to problems.} ReACT with Toolkit outperformed ReACT in all metrics. For instance, G1-Tool's pass rate increases by 9.0\%, and G3-Inst's win rate rises by 4.3\%. These results demonstrate significant performance enhancements. Moreover, the overall performance of ReACT with Toolkit is comparable to AdaPlanner. For example, in G2-Inst, the pass rate of ReACT with Toolkit is only 3.0\% lower than that of AdaPlanner, while in G2-Cat, its win rate is 8.5\% higher. Overall, ReACT with Toolkit significantly improves model performance through tool integration, narrowing the gap with better-performing algorithms and showing higher pass and win rates. 

Similarly, the performance of AdaPlanner significantly improved after integrating the toolkit. Meanwhile, the pass and win rates of AdaPlanner with Toolkit are slightly higher than those of DFSDT. For instance, in G2-Inst, its win rate exceeds that of DFSDT by 4.0\%, with similar improvements observed in other datasets. This indicates that after applying our tool clustering algorithm and using the toolkit as node states in the search process, the initially underperforming AdaPlanner can surpass DFSDT, which doesn't use any classification methods, fully demonstrating the superiority of the clustering algorithm in tool planning.

Our Tool-Planner method derives upon DFSDT by replacing its API node with the toolkit. As seen from the table, the performance of DFSDT improved with the adoption of our clustering algorithm. The win rate surges by 13.5\% in G2-Cat and soares to 92.0\% in G3-Inst. This demonstrates that our innovatively proposed toolkit exhibits significant enhancements in both single and multiple-tool usage scenarios, confirming its effectiveness in practical applications and heralding a new advancement in the field of tool utilization.

\textbf{Impact on different numbers of clusters.} For Tool-Planner, the size of $k$ value in tool clustering has a crucial impact on overall performance. Considering that tools with identical function have closely related tool embeddings, an appropriate number of clusters helps in properly categorizing tools by their functionality. When performing task planning, the model focuses more on the problems that the toolkit can solve rather than the specific details of each API. We set a range of $k$ values and conduct experiments on pass rate and win rate to understand the relationship between configuration size and model performance.

\begin{figure*}[t] 
  \begin{minipage}[b]{0.46\textwidth}
    \centering
    \includegraphics[width=0.98\linewidth]{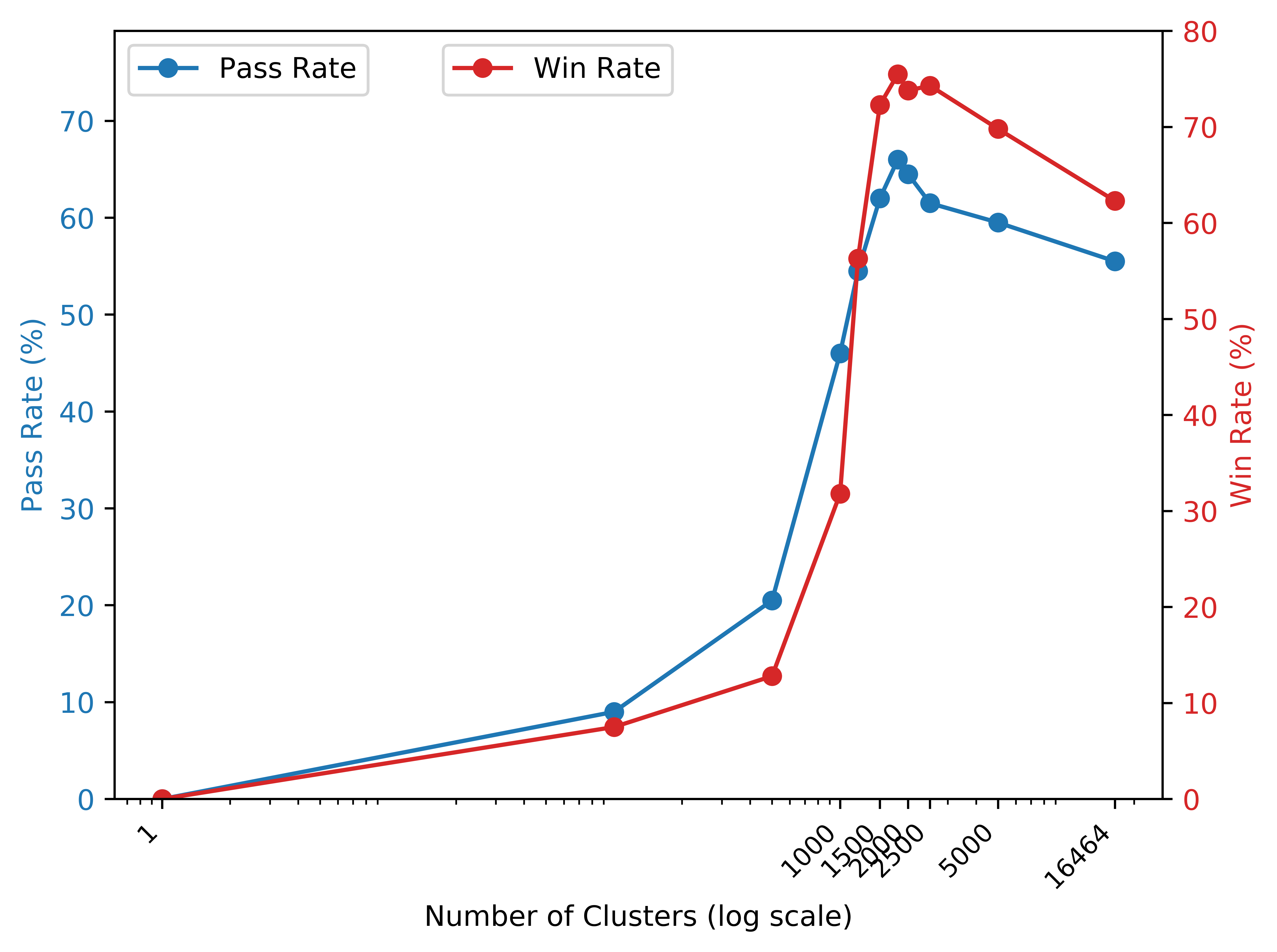}
    \caption{Evaluation of the effects of different numbers of clusters on G1-inst.}
    \label{fig:k}
  \end{minipage}%
  \hfill
   \begin{minipage}[b]{0.46\textwidth}
    \centering
    \includegraphics[width=0.98\linewidth]{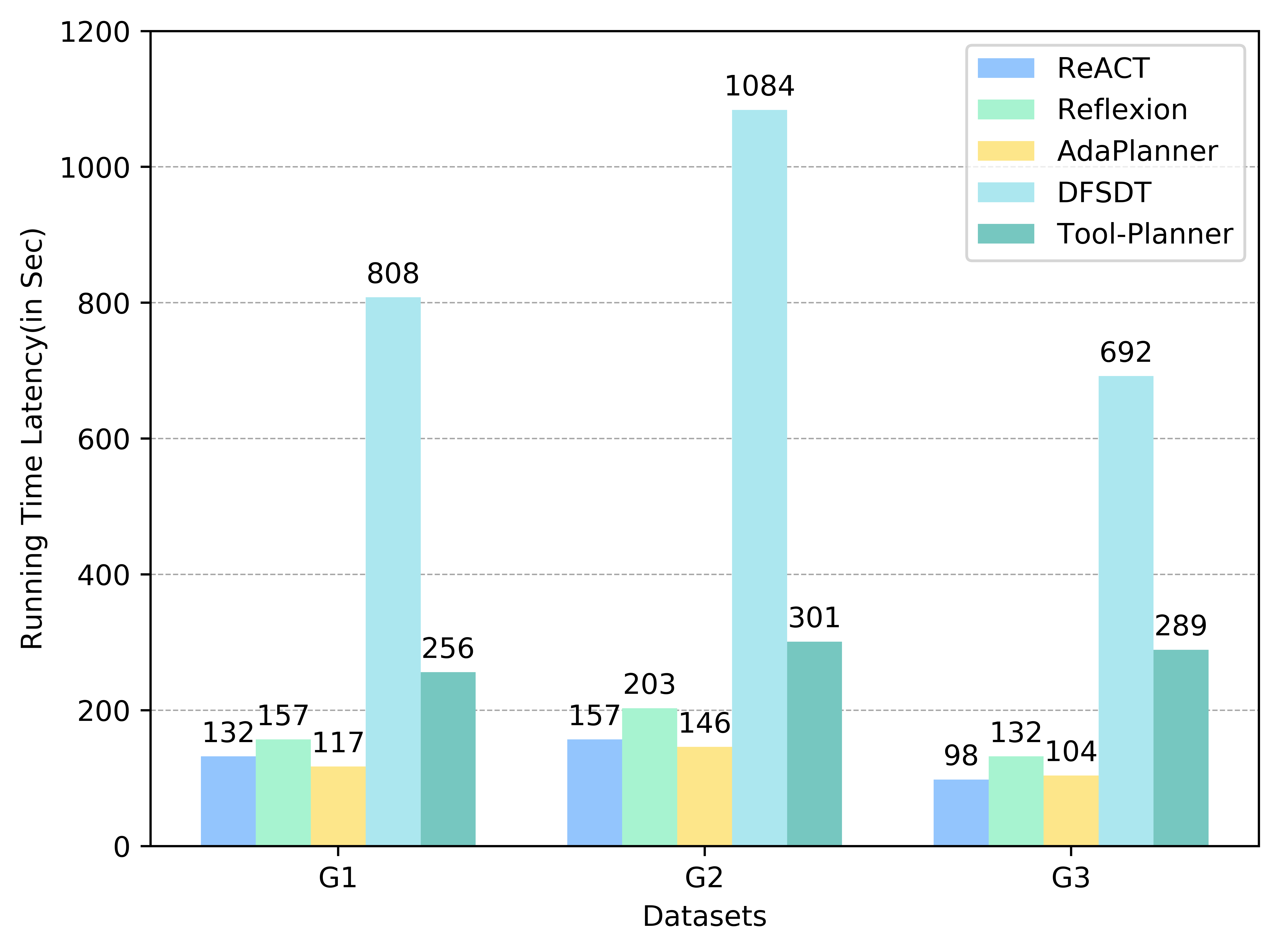}
    \caption{The average running time (sec) of different methods in various datasets.}
    \label{fig:zhu}
  \end{minipage}%
  \hfill
  \vspace{-9pt}
\end{figure*}

On ToolBench, G1-inst achieves the best results when the average size of a cluster is about 9. As shown in Figure~\ref{fig:k}, while $k$ is large, the performance of our framework gradually declined. This indicates that with insufficiently classified tools, the time consumption for solution space search and re-planning is considerable. Additionally, due to insufficient exploration of the solution space, some feasible solutions are not obtained, resulting in a lower overall pass rate and win rate. On the other hand, when $k$ is too small, the clustering effect significantly deteriorates, causing the clustering of different methods to mix together. This leads to incorrect tool calls and the generation of unsolvable next-step information. Therefore, a reasonable number of clusters is crucial for problem-solving. This is closely related to the distribution of tools and datasets.

\begin{wraptable}{r}{0.5\linewidth}
	\centering
 \caption{The pass rate (\%) results of different text clustering models on Tool-Planner.}
 \label{tab:text_emb}
 \vspace{0.1cm}
    \fontsize{9}{9}\selectfont\setlength{\tabcolsep}{0.3em}
\begin{tabular}{llll}
\toprule
\textbf{Model}                 & \textbf{G1-inst.} & \textbf{G2-inst.} & \textbf{G3-inst.} \\ \midrule
RoBERTa-base          & 60.5     & 76.5     & 73.0     \\ 
Contriever            & 61.0     & 78.5     & 76.5     \\ 
\textit{text-embedding-ada-02} & 63.5     & 81.5     & 78.0     \\ 
SimCSE          & \textbf{66.0}     & \textbf{83.5}     & \textbf{83.0}     \\ \bottomrule
\end{tabular}
\end{wraptable}

\textbf{Text embeddings model on tool clustering.} Tool clustering simulates the human process of categorizing tools. In Tool-Planner, the effectiveness of clustering and the similarity of the resulting toolkits are crucial to our method. Tool clustering learns from tool demonstration and documentation, generating sentence embeddings based on their respective functions. In the tool clustering process, we use the SimCSE model to calculate the similarity between tools. To compare the clustering effectiveness of different similarity algorithms and their impact on the final results, we experiment with various similarity algorithms, including different text embedding models like RoBERTa-base \citep{liu2019roberta}, Contriever \citep{izacard2022unsupervised}, and \textit{text-embedding-ada-02}. We evaluate these algorithms based on pass rate metrics.

Table~\ref{tab:text_emb} shows the impact of different text embedding models on the final results. \textbf{SimCSE shows robust ability to generate tool embeddings.} It can be seen that the SimCSE model achieves the best performance among the four text embedding models, indicating that SimCSE can better understand the informational knowledge of tool functions. Meanwhile, the performance of different text embedding models is generally similar, but task-specific embeddings generated through fine-tuning may perform better in more suitable scenarios.

\textbf{Efficiency evaluation.} To comprehensively understand the overhead of different algorithms in practical applications, we compare their execution speeds and test them on various datasets. This comparison not only demonstrates the efficiency differences between the methods but also helps us understand the distinct planning processes and the number of tool invocations for each algorithm. The evaluation results are shown in Figure~\ref{fig:zhu}.

Tool-Planner demonstrates a significant efficiency improvement compared to DFSDT. When an error occurs, Tool-Planner immediately selects another tool with the same function from the toolkit, quickly completing the tool replacement without affecting the original plan. It fully explores the feasible area, attempting alternative paths only after all APIs in the toolkit have been tried. On the other hand, DFSDT tries to find an API when an error occurs. If it determines there are no feasible APIs, it abandons the current state and continues searching. the new planning methods proposed by LLMs are generally more complex, leading to a higher likelihood of tool errors on new planning paths. This repeated plan adjustment not only inadequately explores feasible solutions but also wastes previous computation results. Moreover, we can see that, compared to path search solutions like ReACT and AdaPlanner, the latency of Tool-Planner is only twice as much, while DFSDT's latency is 6-8 times higher. This indicates that Tool-Planner can effectively find feasible tools and select appropriate plans for problem reasoning.


\subsection{Error Analysis}

In this section, we summarize and categorize the issues directly leading to failure or having potential failure risks in each step to conduct a more specific analysis. The types and distribution of failures are shown in Table~\ref{fig:circle}. We also provide examples of each error type in Appendix \ref{error}.

\begin{wrapfigure}{r}{8cm} 
\includegraphics[width=8cm]{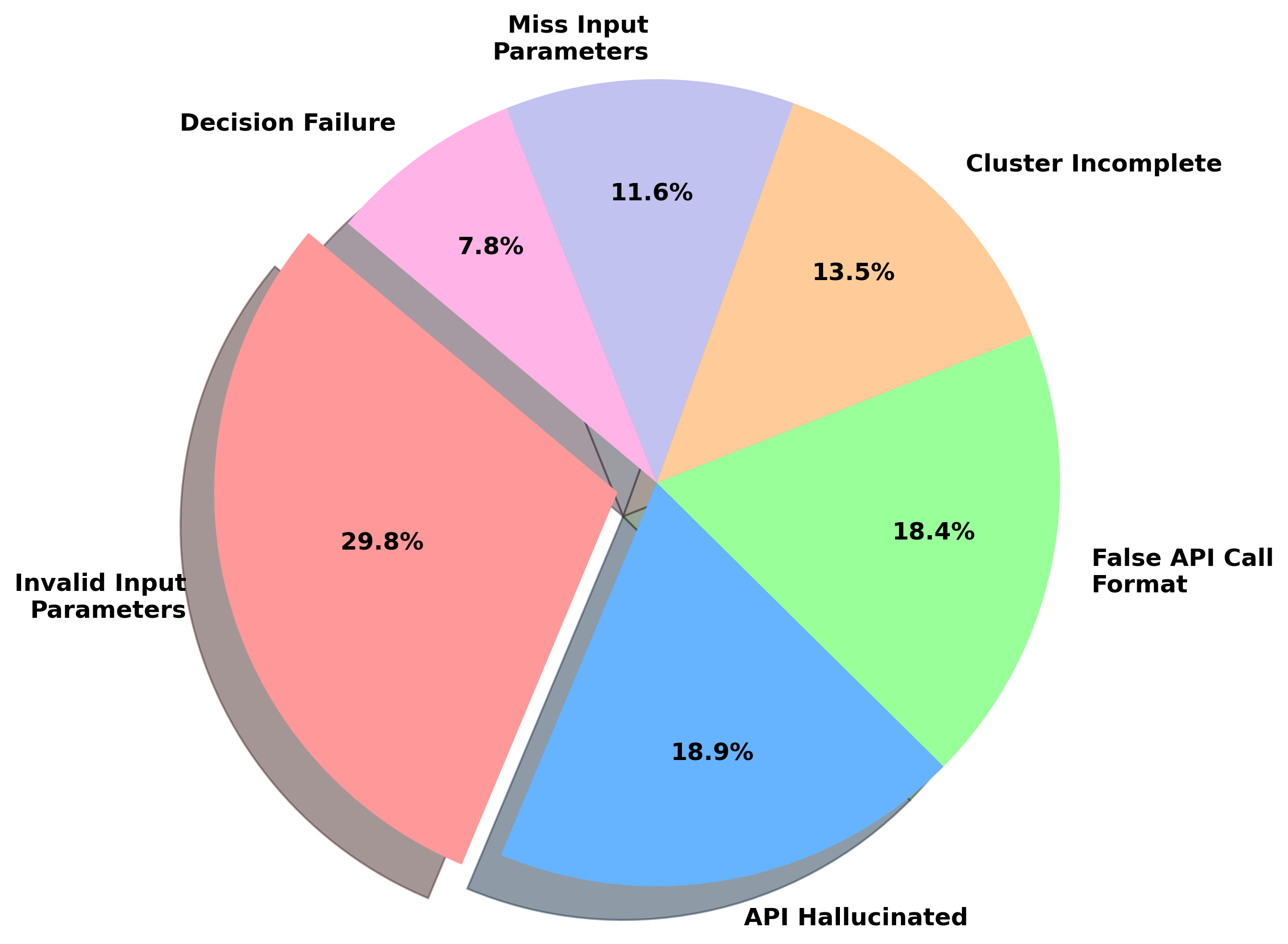}
\caption{Distribution of reasoning errors.}
\label{fig:circle}
\end{wrapfigure}

In the Figure~\ref{fig:circle}, we can see that \textit{Invalid Input Parameters} is the most frequent error type. There are primarily two reasons for this error: \textbf{(1)} When calling the API, the parameters provided to it do not meet the expected content, resulting in invalid input. \textbf{(2)} Users may misunderstand the content of the parameters required. Similar error types include \textit{False API Call Format} and \textit{Miss Input Parameters}, both of which arise during API invocation. Methods to mitigate these errors include providing users with more understandable prompts during the information input stage and stricter validation and filtering of input data during the model inference stage to ensure data integrity and accuracy.

The second most common error type is \textit{API Hallucinated}, which is a prevalent mistake. This occurs when the model attempts to call an API as proposed in the plan but cannot find an API with a matching name. Methods to mitigate the hallucination in LLMs include providing clear and accurate API documentation to enable the model to interpret and use the information correctly, and timely updating the model with information about APIs to avoid hallucination issues caused by API updates.

Furthermore, \textit{Cluster Incomplete} error occurs when the model overlooks some APIs due to incomplete clustering, resulting in the failure to identify APIs that could solve the problem. This type of error occurs due to the newly introduced clustering algorithm in this paper, but the occurrence rate is low, at only \textbf{13.5\%}. Methods to mitigate this error include improving the clustering algorithm to enhance its performance and providing clear and accurate API documentation to facilitate clustering.

\section{Related Work}
\label{others}
\textbf{Task planning with LLMs.} Trained on extensive corpora, LLMs encompass a wealth of common-sense knowledge for task planning~\citep{pallagani2023understanding,sun2023pearl}. Consequently, generative methods have emerged as a hot topic in recent years. In considering the utilization of LLMs, some studies directly generate entire plans without executing them in the environment~\citep{singh2023progprompt,wu2023embodied,liang2023code,lin2023grounded,yang2023foundation,zeng2022socratic}. However, these studies ignore the mechanism to correct decisions, which could result in a chain of errors starting from the initial ones. Reflexion~\citep{shinn2024reflexion} mitigates this issue by requiring LLMs to reflect on past failures. The DFSDT proposed by ToolLLM~\citep{qin2024toolllm} extends Reflexion to a more general method by allowing LLMs to evaluate different reasoning paths and select the most promising one. Our approach creatively utilizes toolkits for plan generation.

\textbf{Tree-based modeling for inference in LLMs.} Most LLM-based agents employ either open-loop or closed-loop systems, relying on linear reasoning or planning structures. To explore multiple branches in the action space, Self-consistency~\citep{wang2022self} samples multiple chains of thought, which can be seen as multiple i.i.d. solution paths in the decision space, and selects the best answer through majority voting. Some works~\citep{yao2024tree,long2023large} propose an alternative solution of chains of thought, called ``tree-of-thought''. These studies focus on reasoning tasks without involving interaction between the internal steps of the tree and the environment. Additionally, RAP~\citep{hao2023reasoning} combines world models with rewards in advanced MCTS search methods. To avoid exhaustive exploration like MCTS, Toolchain*~\citep{zhuang2023toolchain} proposes a method that integrates efficient A* search with the effective reasoning capability of LLM, and Tree-Planner~\citep{hu2023tree} samples different paths once and aggregated them into an action tree. Most methods fail in multi tools scenarios, but our Tool-Planner effectively addresses the issue of tool usage efficiency.

\textbf{LLMs for tool use.} The latest research in language modeling explores the use of external tools to complement the knowledge stored in model weights~\citep{qin2023tool}. This approach allows tasks like precise computation or information retrieval to be offloaded to external modules, such as Python interpreters or search engines~\citep{mialon2023augmented}. These tools retrieve natural language knowledge from additional resources, as demonstrated by WebGPT~\citep{nakano2021webgpt} and ReACT~\citep{yao2022ReACT}, which utilize search APIs to tap into these sources. Other methods, such as Toolformer~\citep{schick2024toolformer}, ART~\citep{paranjape2023art}, ToolkenGPT~\citep{hao2024toolkengpt}, leverage combinations of search APIs, question-answer APIs, machine translation APIs, calculators, and other tools to address various NLP tasks. ChatGPT Plugin\footnote{
https://openai.com/blog/chatgpt-plugins
} and TaskMatrix.AI~\citep{liang2023taskmatrix} show the potential of LLMs integrated with thousands to millions of APIs. LATM~\citep{cai2023large} and CREATOR~\citep{qian2023creator} utilize GPT-4 to create API tools. Our Tool-Planner integrates APIs with the same or similar functions into toolkits, which significantly enhancing the ability to solve sub-problems.

\section{Conclusion}

In this paper, we present Tool-Planner, a framework for task planning based on tool clustering in tool learning. This framework enables flexible adjustments among tools with the same function. When errors occur in task planning, other tools within the same toolkit can be selected to maintain relative consistency of the plan and ensure effective and thorough exploration of the solution space. After all tools within a toolkit have been attempted, we switch to a new toolkit-based task planning to dynamically adjust the planning process. Compared to existing algorithms, our method finds the API to solve the current task more quickly, thus completing the task. Experiments show that our method has a higher pass rate and win rate compared to different baselines. Additionally, the ablation study demonstrates the effectiveness of the toolkit design. We also explore the impact of different clustering numbers and text similarity algorithms on clustering and planning effectiveness. Experimental results further confirm that our method can quickly complete task solutions. We believe this framework will contribute to the long-term development of the tool learning paradigm.

\section{Acknowledgements}

This work was partly supported by the NSFC under No. 62402418 and No. 62102360. This work was also partly supported by the Key R\&D Program of Ningbo under No. 2024Z115.

\bibliography{custom}

\begin{thebibliography}{57}
\providecommand{\natexlab}[1]{#1}
\providecommand{\url}[1]{\texttt{#1}}
\expandafter\ifx\csname urlstyle\endcsname\relax
  \providecommand{\doi}[1]{doi: #1}\else
  \providecommand{\doi}{doi: \begingroup \urlstyle{rm}\Url}\fi

\bibitem[Achiam et~al.(2023)Achiam, Adler, Agarwal, Ahmad, Akkaya, Aleman, Almeida, Altenschmidt, Altman, Anadkat, et~al.]{achiam2023gpt}
Josh Achiam, Steven Adler, Sandhini Agarwal, Lama Ahmad, Ilge Akkaya, Florencia~Leoni Aleman, Diogo Almeida, Janko Altenschmidt, Sam Altman, Shyamal Anadkat, et~al.
\newblock Gpt-4 technical report.
\newblock \emph{arXiv preprint arXiv:2303.08774}, 2023.

\bibitem[Ahmad et~al.(2023)Ahmad, Kaiser, and Rahim]{ahmad2023hallucinations}
Zakia Ahmad, Wahid Kaiser, and Sifatur Rahim.
\newblock Hallucinations in chatgpt: an unreliable tool for learning.
\newblock \emph{Rupkatha Journal on Interdisciplinary Studies in Humanities}, 15\penalty0 (4):\penalty0 1--18, 2023.

\bibitem[Arthur \& Vassilvitskii(2007)Arthur and Vassilvitskii]{arthur2007k}
David Arthur and Sergei Vassilvitskii.
\newblock k-means++ the advantages of careful seeding.
\newblock In \emph{Proceedings of the eighteenth annual ACM-SIAM symposium on Discrete algorithms}, pp.\  1027--1035, 2007.

\bibitem[Bang et~al.(2023)Bang, Cahyawijaya, Lee, Dai, Su, Wilie, Lovenia, Ji, Yu, Chung, et~al.]{bang2023multitask}
Yejin Bang, Samuel Cahyawijaya, Nayeon Lee, Wenliang Dai, Dan Su, Bryan Wilie, Holy Lovenia, Ziwei Ji, Tiezheng Yu, Willy Chung, et~al.
\newblock A multitask, multilingual, multimodal evaluation of chatgpt on reasoning, hallucination, and interactivity.
\newblock \emph{arXiv preprint arXiv:2302.04023}, 2023.

\bibitem[Brown et~al.(2020)Brown, Mann, Ryder, Subbiah, Kaplan, Dhariwal, Neelakantan, Shyam, Sastry, Askell, et~al.]{GPT3}
Tom Brown, Benjamin Mann, Nick Ryder, Melanie Subbiah, Jared~D Kaplan, Prafulla Dhariwal, Arvind Neelakantan, Pranav Shyam, Girish Sastry, Amanda Askell, et~al.
\newblock Language models are few-shot learners.
\newblock \emph{Advances in Neural Information Processing Systems}, 33:\penalty0 1877--1901, 2020.

\bibitem[Cai et~al.(2023)Cai, Wang, Ma, Chen, and Zhou]{cai2023large}
Tianle Cai, Xuezhi Wang, Tengyu Ma, Xinyun Chen, and Denny Zhou.
\newblock Large language models as tool makers.
\newblock \emph{arXiv preprint arXiv:2305.17126}, 2023.

\bibitem[Chowdhery et~al.(2023)Chowdhery, Narang, Devlin, Bosma, Mishra, Roberts, Barham, Chung, Sutton, Gehrmann, et~al.]{chowdhery2023palm}
Aakanksha Chowdhery, Sharan Narang, Jacob Devlin, Maarten Bosma, Gaurav Mishra, Adam Roberts, Paul Barham, Hyung~Won Chung, Charles Sutton, Sebastian Gehrmann, et~al.
\newblock Palm: Scaling language modeling with pathways.
\newblock \emph{Journal of Machine Learning Research}, 24\penalty0 (240):\penalty0 1--113, 2023.

\bibitem[Du et~al.(2024)Du, Wei, and Zhang]{du2024anytool}
Yu~Du, Fangyun Wei, and Hongyang Zhang.
\newblock Anytool: Self-reflective, hierarchical agents for large-scale api calls.
\newblock \emph{arXiv preprint arXiv:2402.04253}, 2024.

\bibitem[Fu et~al.(2023)Fu, Peng, Sabharwal, Clark, and Khot]{fu2023complexitybased}
Yao Fu, Hao Peng, Ashish Sabharwal, Peter Clark, and Tushar Khot.
\newblock Complexity-based prompting for multi-step reasoning.
\newblock In \emph{The Eleventh International Conference on Learning Representations}, 2023.
\newblock URL \url{https://openreview.net/forum?id=yf1icZHC-l9}.

\bibitem[Gao et~al.(2021)Gao, Yao, and Chen]{gao2021simcse}
Tianyu Gao, Xingcheng Yao, and Danqi Chen.
\newblock Simcse: Simple contrastive learning of sentence embeddings.
\newblock In \emph{Proceedings of the 2021 Conference on Empirical Methods in Natural Language Processing}, pp.\  6894--6910, 2021.

\bibitem[Gou et~al.(2024{\natexlab{a}})Gou, Shao, Gong, yelong shen, Yang, Duan, and Chen]{gou2024critic}
Zhibin Gou, Zhihong Shao, Yeyun Gong, yelong shen, Yujiu Yang, Nan Duan, and Weizhu Chen.
\newblock {CRITIC}: Large language models can self-correct with tool-interactive critiquing.
\newblock In \emph{The Twelfth International Conference on Learning Representations}, 2024{\natexlab{a}}.
\newblock URL \url{https://openreview.net/forum?id=Sx038qxjek}.

\bibitem[Gou et~al.(2024{\natexlab{b}})Gou, Shao, Gong, yelong shen, Yang, Huang, Duan, and Chen]{gou2024tora}
Zhibin Gou, Zhihong Shao, Yeyun Gong, yelong shen, Yujiu Yang, Minlie Huang, Nan Duan, and Weizhu Chen.
\newblock To{RA}: A tool-integrated reasoning agent for mathematical problem solving.
\newblock In \emph{The Twelfth International Conference on Learning Representations}, 2024{\natexlab{b}}.
\newblock URL \url{https://openreview.net/forum?id=Ep0TtjVoap}.

\bibitem[Guerreiro et~al.(2023)Guerreiro, Alves, Waldendorf, Haddow, Birch, Colombo, and Martins]{guerreiro2023hallucinations}
Nuno~M Guerreiro, Duarte~M Alves, Jonas Waldendorf, Barry Haddow, Alexandra Birch, Pierre Colombo, and Andr{\'e}~FT Martins.
\newblock Hallucinations in large multilingual translation models.
\newblock \emph{Transactions of the Association for Computational Linguistics}, 11:\penalty0 1500--1517, 2023.

\bibitem[Hao et~al.(2023{\natexlab{a}})Hao, Gu, Ma, Hong, Wang, Wang, and Hu]{hao2023reasoning}
Shibo Hao, Yi~Gu, Haodi Ma, Joshua~Jiahua Hong, Zhen Wang, Daisy~Zhe Wang, and Zhiting Hu.
\newblock Reasoning with language model is planning with world model.
\newblock \emph{arXiv preprint arXiv:2305.14992}, 2023{\natexlab{a}}.

\bibitem[Hao et~al.(2023{\natexlab{b}})Hao, Liu, Wang, and Hu]{hao2024toolkengpt}
Shibo Hao, Tianyang Liu, Zhen Wang, and Zhiting Hu.
\newblock Toolkengpt: Augmenting frozen language models with massive tools via tool embeddings.
\newblock \emph{Advances in Neural Information Processing Systems}, 36, 2023{\natexlab{b}}.

\bibitem[Hu et~al.(2023)Hu, Mu, Yu, Ding, Wu, Shao, Chen, Wang, Qiao, and Luo]{hu2023tree}
Mengkang Hu, Yao Mu, Xinmiao~Chelsey Yu, Mingyu Ding, Shiguang Wu, Wenqi Shao, Qiguang Chen, Bin Wang, Yu~Qiao, and Ping Luo.
\newblock Tree-planner: Efficient close-loop task planning with large language models.
\newblock In \emph{The Twelfth International Conference on Learning Representations}, 2023.

\bibitem[Izacard et~al.(2022)Izacard, Caron, Hosseini, Riedel, Bojanowski, Joulin, and Grave]{izacard2022unsupervised}
Gautier Izacard, Mathilde Caron, Lucas Hosseini, Sebastian Riedel, Piotr Bojanowski, Armand Joulin, and Edouard Grave.
\newblock Unsupervised dense information retrieval with contrastive learning.
\newblock \emph{Transactions on Machine Learning Research}, 2022.
\newblock ISSN 2835-8856.
\newblock URL \url{https://openreview.net/forum?id=jKN1pXi7b0}.

\bibitem[Lee et~al.(2024)Lee, Smith, Woodhead, and Lan]{lee2024math}
Jaewook Lee, Digory Smith, Simon Woodhead, and Andrew Lan.
\newblock Math multiple choice question generation via human-large language model collaboration.
\newblock \emph{arXiv preprint arXiv:2405.00864}, 2024.

\bibitem[Liang et~al.(2023{\natexlab{a}})Liang, Huang, Xia, Xu, Hausman, Ichter, Florence, and Zeng]{liang2023code}
Jacky Liang, Wenlong Huang, Fei Xia, Peng Xu, Karol Hausman, Brian Ichter, Pete Florence, and Andy Zeng.
\newblock Code as policies: Language model programs for embodied control.
\newblock In \emph{2023 IEEE International Conference on Robotics and Automation (ICRA)}, pp.\  9493--9500. IEEE, 2023{\natexlab{a}}.

\bibitem[Liang et~al.(2023{\natexlab{b}})Liang, Wu, Song, Wu, Xia, Liu, Ou, Lu, Ji, Mao, et~al.]{liang2023taskmatrix}
Yaobo Liang, Chenfei Wu, Ting Song, Wenshan Wu, Yan Xia, Yu~Liu, Yang Ou, Shuai Lu, Lei Ji, Shaoguang Mao, et~al.
\newblock Taskmatrix. ai: Completing tasks by connecting foundation models with millions of apis.
\newblock \emph{arXiv preprint arXiv:2303.16434}, 2023{\natexlab{b}}.

\bibitem[Lin et~al.(2023)Lin, Huang, Liu, Gu, Sommerer, and Ren]{lin2023grounded}
Bill~Yuchen Lin, Chengsong Huang, Qian Liu, Wenda Gu, Sam Sommerer, and Xiang Ren.
\newblock On grounded planning for embodied tasks with language models.
\newblock In \emph{Proceedings of the AAAI Conference on Artificial Intelligence}, volume~37, pp.\  13192--13200, 2023.

\bibitem[Liu et~al.(2024{\natexlab{a}})Liu, Peng, Du, Yin, Liu, and Zhang]{liu2024era}
Yanming Liu, Xinyue Peng, Tianyu Du, Jianwei Yin, Weihao Liu, and Xuhong Zhang.
\newblock Era-cot: Improving chain-of-thought through entity relationship analysis.
\newblock \emph{arXiv preprint arXiv:2403.06932}, 2024{\natexlab{a}}.

\bibitem[Liu et~al.(2024{\natexlab{b}})Liu, Peng, Zhang, Liu, Yin, Cao, and Du]{liu2024ra}
Yanming Liu, Xinyue Peng, Xuhong Zhang, Weihao Liu, Jianwei Yin, Jiannan Cao, and Tianyu Du.
\newblock Ra-isf: Learning to answer and understand from retrieval augmentation via iterative self-feedback.
\newblock \emph{arXiv preprint arXiv:2403.06840}, 2024{\natexlab{b}}.

\bibitem[Liu et~al.(2019)Liu, Ott, Goyal, Du, Joshi, Chen, Levy, Lewis, Zettlemoyer, and Stoyanov]{liu2019roberta}
Yinhan Liu, Myle Ott, Naman Goyal, Jingfei Du, Mandar Joshi, Danqi Chen, Omer Levy, Mike Lewis, Luke Zettlemoyer, and Veselin Stoyanov.
\newblock Roberta: A robustly optimized bert pretraining approach.
\newblock \emph{arXiv preprint arXiv:1907.11692}, 2019.

\bibitem[Long(2023)]{long2023large}
Jieyi Long.
\newblock Large language model guided tree-of-thought.
\newblock \emph{arXiv preprint arXiv:2305.08291}, 2023.

\bibitem[Lu et~al.(2024)Lu, Peng, Cheng, Galley, Chang, Wu, Zhu, and Gao]{lu2024chameleon}
Pan Lu, Baolin Peng, Hao Cheng, Michel Galley, Kai-Wei Chang, Ying~Nian Wu, Song-Chun Zhu, and Jianfeng Gao.
\newblock Chameleon: Plug-and-play compositional reasoning with large language models.
\newblock \emph{Advances in Neural Information Processing Systems}, 36, 2024.

\bibitem[Mialon et~al.(2023)Mialon, Dess{\`\i}, Lomeli, Nalmpantis, Pasunuru, Raileanu, Rozi{\`e}re, Schick, Dwivedi-Yu, Celikyilmaz, et~al.]{mialon2023augmented}
Gr{\'e}goire Mialon, Roberto Dess{\`\i}, Maria Lomeli, Christoforos Nalmpantis, Ram Pasunuru, Roberta Raileanu, Baptiste Rozi{\`e}re, Timo Schick, Jane Dwivedi-Yu, Asli Celikyilmaz, et~al.
\newblock Augmented language models: a survey.
\newblock \emph{arXiv preprint arXiv:2302.07842}, 2023.

\bibitem[Nakano et~al.(2021)Nakano, Hilton, Balaji, Wu, Ouyang, Kim, Hesse, Jain, Kosaraju, Saunders, et~al.]{nakano2021webgpt}
Reiichiro Nakano, Jacob Hilton, Suchir Balaji, Jeff Wu, Long Ouyang, Christina Kim, Christopher Hesse, Shantanu Jain, Vineet Kosaraju, William Saunders, et~al.
\newblock Webgpt: Browser-assisted question-answering with human feedback.
\newblock \emph{arXiv preprint arXiv:2112.09332}, 2021.

\bibitem[Ouyang et~al.(2022)Ouyang, Wu, Jiang, Almeida, Wainwright, Mishkin, Zhang, Agarwal, Slama, Ray, et~al.]{ouyang2022training}
Long Ouyang, Jeffrey Wu, Xu~Jiang, Diogo Almeida, Carroll Wainwright, Pamela Mishkin, Chong Zhang, Sandhini Agarwal, Katarina Slama, Alex Ray, et~al.
\newblock Training language models to follow instructions with human feedback.
\newblock \emph{Advances in neural information processing systems}, 35:\penalty0 27730--27744, 2022.

\bibitem[Pallagani et~al.(2023)Pallagani, Muppasani, Murugesan, Rossi, Srivastava, Horesh, Fabiano, and Loreggia]{pallagani2023understanding}
Vishal Pallagani, Bharath Muppasani, Keerthiram Murugesan, Francesca Rossi, Biplav Srivastava, Lior Horesh, Francesco Fabiano, and Andrea Loreggia.
\newblock Understanding the capabilities of large language models for automated planning.
\newblock \emph{arXiv preprint arXiv:2305.16151}, 2023.

\bibitem[Paranjape et~al.(2023)Paranjape, Lundberg, Singh, Hajishirzi, Zettlemoyer, and Ribeiro]{paranjape2023art}
Bhargavi Paranjape, Scott Lundberg, Sameer Singh, Hannaneh Hajishirzi, Luke Zettlemoyer, and Marco~Tulio Ribeiro.
\newblock Art: Automatic multi-step reasoning and tool-use for large language models.
\newblock \emph{arXiv preprint arXiv:2303.09014}, 2023.

\bibitem[Patil et~al.(2023)Patil, Zhang, Wang, and Gonzalez]{patil2023gorilla}
Shishir~G. Patil, Tianjun Zhang, Xin Wang, and Joseph~E. Gonzalez.
\newblock Gorilla: Large language model connected with massive apis.
\newblock \emph{arXiv preprint arXiv:2305.15334}, 2023.

\bibitem[Qian et~al.(2023)Qian, Han, Fung, Qin, Liu, and Ji]{qian2023creator}
Cheng Qian, Chi Han, Yi~R Fung, Yujia Qin, Zhiyuan Liu, and Heng Ji.
\newblock Creator: Disentangling abstract and concrete reasonings of large language models through tool creation.
\newblock \emph{arXiv preprint arXiv:2305.14318}, 2023.

\bibitem[Qin et~al.(2023)Qin, Hu, Lin, Chen, Ding, Cui, Zeng, Huang, Xiao, Han, et~al.]{qin2023tool}
Yujia Qin, Shengding Hu, Yankai Lin, Weize Chen, Ning Ding, Ganqu Cui, Zheni Zeng, Yufei Huang, Chaojun Xiao, Chi Han, et~al.
\newblock Tool learning with foundation models.
\newblock \emph{arXiv preprint arXiv:2304.08354}, 2023.

\bibitem[Qin et~al.(2024)Qin, Liang, Ye, Zhu, Yan, Lu, Lin, Cong, Tang, Qian, Zhao, Hong, Tian, Xie, Zhou, Gerstein, dahai li, Liu, and Sun]{qin2024toolllm}
Yujia Qin, Shihao Liang, Yining Ye, Kunlun Zhu, Lan Yan, Yaxi Lu, Yankai Lin, Xin Cong, Xiangru Tang, Bill Qian, Sihan Zhao, Lauren Hong, Runchu Tian, Ruobing Xie, Jie Zhou, Mark Gerstein, dahai li, Zhiyuan Liu, and Maosong Sun.
\newblock Tool{LLM}: Facilitating large language models to master 16000+ real-world {API}s.
\newblock In \emph{The Twelfth International Conference on Learning Representations}, 2024.
\newblock URL \url{https://openreview.net/forum?id=dHng2O0Jjr}.

\bibitem[Rapid(2023)]{rapidapi}
Rapid.
\newblock Rapid api, 2023.
\newblock \url{https://rapidapi.com/}.

\bibitem[Schick et~al.(2023)Schick, Dwivedi-Yu, Dess{\`\i}, Raileanu, Lomeli, Hambro, Zettlemoyer, Cancedda, and Scialom]{schick2024toolformer}
Timo Schick, Jane Dwivedi-Yu, Roberto Dess{\`\i}, Roberta Raileanu, Maria Lomeli, Eric Hambro, Luke Zettlemoyer, Nicola Cancedda, and Thomas Scialom.
\newblock Toolformer: Language models can teach themselves to use tools.
\newblock \emph{Advances in Neural Information Processing Systems}, 36, 2023.

\bibitem[Shen et~al.(2024)Shen, Song, Tan, Li, Lu, and Zhuang]{shen2024hugginggpt}
Yongliang Shen, Kaitao Song, Xu~Tan, Dongsheng Li, Weiming Lu, and Yueting Zhuang.
\newblock Hugginggpt: Solving ai tasks with chatgpt and its friends in hugging face.
\newblock \emph{Advances in Neural Information Processing Systems}, 36, 2024.

\bibitem[Shinn et~al.(2023)Shinn, Cassano, Gopinath, Narasimhan, and Yao]{shinn2024reflexion}
Noah Shinn, Federico Cassano, Ashwin Gopinath, Karthik Narasimhan, and Shunyu Yao.
\newblock Reflexion: Language agents with verbal reinforcement learning.
\newblock \emph{Advances in Neural Information Processing Systems}, 36, 2023.

\bibitem[Singh et~al.(2023)Singh, Blukis, Mousavian, Goyal, Xu, Tremblay, Fox, Thomason, and Garg]{singh2023progprompt}
Ishika Singh, Valts Blukis, Arsalan Mousavian, Ankit Goyal, Danfei Xu, Jonathan Tremblay, Dieter Fox, Jesse Thomason, and Animesh Garg.
\newblock Progprompt: Generating situated robot task plans using large language models.
\newblock In \emph{2023 IEEE International Conference on Robotics and Automation (ICRA)}, pp.\  11523--11530. IEEE, 2023.

\bibitem[Song et~al.(2023)Song, Xiong, Zhu, Li, Wang, Tian, and Li]{song2023restgpt}
Yifan Song, Weimin Xiong, Dawei Zhu, Cheng Li, Ke~Wang, Ye~Tian, and Sujian Li.
\newblock Restgpt: Connecting large language models with real-world applications via restful apis.
\newblock \emph{arXiv preprint arXiv:2306.06624}, 2023.

\bibitem[Sun et~al.(2023{\natexlab{a}})Sun, Zhuang, Kong, Dai, and Zhang]{sun2024adaplanner}
Haotian Sun, Yuchen Zhuang, Lingkai Kong, Bo~Dai, and Chao Zhang.
\newblock Adaplanner: Adaptive planning from feedback with language models.
\newblock \emph{Advances in Neural Information Processing Systems}, 36, 2023{\natexlab{a}}.

\bibitem[Sun et~al.(2023{\natexlab{b}})Sun, Liu, Wang, Zhu, and Iyyer]{sun2023pearl}
Simeng Sun, Yang Liu, Shuohang Wang, Chenguang Zhu, and Mohit Iyyer.
\newblock Pearl: Prompting large language models to plan and execute actions over long documents.
\newblock \emph{arXiv preprint arXiv:2305.14564}, 2023{\natexlab{b}}.

\bibitem[Tipirneni et~al.(2024)Tipirneni, Adkathimar, Choudhary, Hiranandani, Amjad, Ioannidis, Yuan, and Reddy]{tipirneni2024context}
Sindhu Tipirneni, Ravinarayana Adkathimar, Nurendra Choudhary, Gaurush Hiranandani, Rana~Ali Amjad, Vassilis~N Ioannidis, Changhe Yuan, and Chandan~K Reddy.
\newblock Context-aware clustering using large language models.
\newblock \emph{arXiv preprint arXiv:2405.00988}, 2024.

\bibitem[Touvron et~al.(2023)Touvron, Martin, Stone, Albert, Almahairi, Babaei, Bashlykov, Batra, Bhargava, Bhosale, et~al.]{touvron2023llama}
Hugo Touvron, Louis Martin, Kevin Stone, Peter Albert, Amjad Almahairi, Yasmine Babaei, Nikolay Bashlykov, Soumya Batra, Prajjwal Bhargava, Shruti Bhosale, et~al.
\newblock Llama 2: Open foundation and fine-tuned chat models.
\newblock \emph{arXiv preprint arXiv:2307.09288}, 2023.

\bibitem[Viswanathan et~al.(2024)Viswanathan, Gashteovski, Gashteovski, Lawrence, Wu, and Neubig]{viswanathan2024large}
Vijay Viswanathan, Kiril Gashteovski, Kiril Gashteovski, Carolin Lawrence, Tongshuang Wu, and Graham Neubig.
\newblock Large language models enable few-shot clustering.
\newblock \emph{Transactions of the Association for Computational Linguistics}, 12:\penalty0 321--333, 2024.

\bibitem[Wang et~al.(2022)Wang, Wei, Schuurmans, Le, Chi, Narang, Chowdhery, and Zhou]{wang2022self}
Xuezhi Wang, Jason Wei, Dale Schuurmans, Quoc Le, Ed~Chi, Sharan Narang, Aakanksha Chowdhery, and Denny Zhou.
\newblock Self-consistency improves chain of thought reasoning in language models.
\newblock \emph{arXiv preprint arXiv:2203.11171}, 2022.

\bibitem[Wei et~al.(2022)Wei, Wang, Schuurmans, Bosma, Xia, Chi, Le, Zhou, et~al.]{wei2022chain}
Jason Wei, Xuezhi Wang, Dale Schuurmans, Maarten Bosma, Fei Xia, Ed~Chi, Quoc~V Le, Denny Zhou, et~al.
\newblock Chain-of-thought prompting elicits reasoning in large language models.
\newblock \emph{Advances in Neural Information Processing Systems}, 35:\penalty0 24824--24837, 2022.

\bibitem[Wu et~al.(2023)Wu, Wang, Xu, Lu, and Yan]{wu2023embodied}
Zhenyu Wu, Ziwei Wang, Xiuwei Xu, Jiwen Lu, and Haibin Yan.
\newblock Embodied task planning with large language models.
\newblock \emph{arXiv preprint arXiv:2307.01848}, 2023.

\bibitem[Yang et~al.(2023)Yang, Nachum, Du, Wei, Abbeel, and Schuurmans]{yang2023foundation}
Sherry Yang, Ofir Nachum, Yilun Du, Jason Wei, Pieter Abbeel, and Dale Schuurmans.
\newblock Foundation models for decision making: Problems, methods, and opportunities.
\newblock \emph{arXiv preprint arXiv:2303.04129}, 2023.

\bibitem[Yao et~al.(2022)Yao, Zhao, Yu, Du, Shafran, Narasimhan, and Cao]{yao2022ReACT}
Shunyu Yao, Jeffrey Zhao, Dian Yu, Nan Du, Izhak Shafran, Karthik~R Narasimhan, and Yuan Cao.
\newblock React: Synergizing reasoning and acting in language models.
\newblock In \emph{The Eleventh International Conference on Learning Representations}, 2022.

\bibitem[Yao et~al.(2024)Yao, Yu, Zhao, Shafran, Griffiths, Cao, and Narasimhan]{yao2024tree}
Shunyu Yao, Dian Yu, Jeffrey Zhao, Izhak Shafran, Tom Griffiths, Yuan Cao, and Karthik Narasimhan.
\newblock Tree of thoughts: Deliberate problem solving with large language models.
\newblock \emph{Advances in Neural Information Processing Systems}, 36, 2024.

\bibitem[Zeng et~al.(2022)Zeng, Attarian, Ichter, Choromanski, Wong, Welker, Tombari, Purohit, Ryoo, Sindhwani, et~al.]{zeng2022socratic}
Andy Zeng, Maria Attarian, Brian Ichter, Krzysztof Choromanski, Adrian Wong, Stefan Welker, Federico Tombari, Aveek Purohit, Michael Ryoo, Vikas Sindhwani, et~al.
\newblock Socratic models: Composing zero-shot multimodal reasoning with language.
\newblock \emph{arXiv preprint arXiv:2204.00598}, 2022.

\bibitem[Zeng et~al.(2023)Zeng, Liu, Du, Wang, Lai, Ding, Yang, Xu, Zheng, Xia, Tam, Ma, Xue, Zhai, Chen, Liu, Zhang, Dong, and Tang]{Zeng2023GLM}
Aohan Zeng, Xiao Liu, Zhengxiao Du, Zihan Wang, Hanyu Lai, Ming Ding, Zhuoyi Yang, Yifan Xu, Wendi Zheng, Xiao Xia, Weng~Lam Tam, Zixuan Ma, Yufei Xue, Jidong Zhai, Wenguang Chen, Zhiyuan Liu, Peng Zhang, Yuxiao Dong, and Jie Tang.
\newblock Glm-130b: An {Open} {Bilingual} {Pre}-trained {Model}.
\newblock In \emph{The Eleventh International Conference on Learning Representations}, 2023.

\bibitem[Zhang et~al.(2022)Zhang, Roller, Goyal, Artetxe, Chen, Chen, Dewan, Diab, Li, Lin, Mihaylov, Ott, Shleifer, Shuster, Simig, Koura, Sridhar, Wang, and Zettlemoyer]{zhang2022opt}
Susan Zhang, Stephen Roller, Naman Goyal, Mikel Artetxe, Moya Chen, Shuohui Chen, Christopher Dewan, Mona Diab, Xian Li, Xi~Victoria Lin, Todor Mihaylov, Myle Ott, Sam Shleifer, Kurt Shuster, Daniel Simig, Punit~Singh Koura, Anjali Sridhar, Tianlu Wang, and Luke Zettlemoyer.
\newblock Opt: Open {Pre}-trained {Transformer} {Language} {Models}.
\newblock \emph{ArXiv}, abs/2205.01068, 2022.

\bibitem[Zhang et~al.(2023)Zhang, Wang, and Shang]{zhang2023clusterllm}
Yuwei Zhang, Zihan Wang, and Jingbo Shang.
\newblock Clusterllm: Large language models as a guide for text clustering.
\newblock In \emph{Proceedings of the 2023 Conference on Empirical Methods in Natural Language Processing}, pp.\  13903--13920, 2023.

\bibitem[Zhuang et~al.(2023)Zhuang, Chen, Yu, Mitra, Bursztyn, Rossi, Sarkhel, and Zhang]{zhuang2023toolchain}
Yuchen Zhuang, Xiang Chen, Tong Yu, Saayan Mitra, Victor Bursztyn, Ryan~A Rossi, Somdeb Sarkhel, and Chao Zhang.
\newblock Toolchain*: Efficient action space navigation in large language models with a* search.
\newblock In \emph{The Twelfth International Conference on Learning Representations}, 2023.

\end{thebibliography}
\bibliographystyle{iclr2025_conference}

\appendix

\section{Broader Impact and Limitations}

\textbf{Broader Impact.} Tool-Planner innovatively integrates toolkits, achieving efficient search in the solution space for task planning. The Tool-Planner paradigm not only performs well on datasets like ToolBench but can also be applied to complex real-world API task scenarios. This approach allows us to integrate APIs of different categories and information types by placing the same type of APIs into a toolkit, enabling multiple attempts with similar tools when addressing practical problems without extensive exploration across a wide solution space.

\textbf{Limitation.} Tool-Planner has some limitations. Firstly, it heavily relies on clustering effectiveness. When there are significant functional differences within a cluster, the model may fail to find the appropriate tool for reasoning. This necessitates setting an appropriate cluster size during clustering to merge tools with similar functions. Secondly, there is still room to explore better tool invocation schemes. In our scenario, compared to linear invocation methods like ReACT \cite{yao2022ReACT}, Tool-Planner still has twice the time delay. We hope to further study methods for tool selection within clusters in the future.

\vspace{0.2cm}
\begin{algorithm}[ht]
\caption{Tool-Planner Exploration Search}\label{alg:remasker}
\textbf{Input:} Query: $x_0$; API docs: $\gD_\gT$; Toolkits functionality: $\{m_{\gT_i}\}_{i=1}^{K}$, Planning length: $s$, Toolkit planning: $\gP_\gT = \{P_i\}_{i=1}^{s} $ \\
\textbf{Output:} Intermediate reasoning step: $\gX = \{x_i\}_{i=1}^{s}$;
\begin{algorithmic}[1]
\State $P_{now} \gets P_1$; $\gX \gets \{x_0\}$; $l \gets 1$; $E \gets \emptyset$;

\While{$l$ is not higher to $s$} \Comment{// Finding a Suitable Answer}
\State $c_{now} \gets \emptyset$
\State $d_{now} \gets \emptyset$
\For{$d_l \in \gD_{\gT_{P_{now}}}$} \Comment{// Task Planning within Same Toolkits}
\State $param \gets $ LLMs$(x_{l-1}, d_l)$;
\State $c_{l} \gets F_{d_l}(param)$; \Comment{// Making Function Call}
\If{$c_l$ is not error message}
\State $c_{now} = c_l$; \Comment{// Fetching Valid Tool}
\State Break Loop;   
\EndIf
\EndFor
\If{$c_{now}$ is not $\emptyset$}
\State $x_{l} \gets $ LLMs$(\gX, \{c_i\}_{i=1}^l, d_{now})$; \Comment{// Generate the Intermediate State Result} 
\State Add  $x_l$  to  $\gX$  under $x_{l-1}$;
\State $l \gets l + 1$;
\EndIf
\If{$c_{now}$ is $\emptyset$} \Comment{// Task Replanning across Toolkits}
\State $\gP'_\gT \gets $ LLMs$(\gX, \{c_i\}_{i=1}^l, \gP_\gT, E)$;
\State Add  $\gP_\gT$  to  $E$;
\State $l \gets LCA(\gP'_\gT, \gP_\gT)$; \Comment{// Restart from Lowest Common Ancestor Node}
\EndIf
\State $P_{now} \gets P_{l}$;
\EndWhile
\State \textbf{return} $\gX = \{x_i\}_{i=1}^{s}$.
\end{algorithmic}
\label{alg:tool}
\end{algorithm}

\section{Implementation Details}

\label{title:B}

\textbf{Dataset.} ToolBench \cite{qin2024toolllm} serves as a benchmark designed to evaluate the API calling capabilities of agents. The ToolBench team gathered 16,464 real-world APIs from RapidAPI \cite{rapidapi} and compiled multiple execution traces for use as a training corpus. This is the only large enough benchmark that contains enough APIs to shows the ability of tool clustering and simulate the real world APIs usage.

The ToolBench test set is categorized into six distinct groups: G1-instruction, G1-tool, G1-category, G2-instruction, G2-category, and G3-instruction. Groups labeled with ``instruction'' include test instructions that utilize tools from the training set, thereby representing in-domain test data. In contrast, groups labeled with ``tool'' or ``category'' feature test instructions that do not use tools from the training set, represent out-of-domain test data. Each group consists of 100 user instructions, totaling 400 instructions for the in-domain test set and 200 instructions for the out-of-domain test set.

\textbf{Environment.} In tool learning, we primarily rely on a series of API function calls to complete planning task processing. In this process, we use toolsets as nodes in the dynamic search tree of tool learning. The SimCSE \cite{gao2021simcse} model we adopt is a pre-trained supervised fine-tuning model based on RoBERTa-base \cite{liu2019roberta}. We generate corresponding explanatory information for each API using the prompts provided in the appendix and embed this information into the KG class. We utilize the Kmeans++ \cite{arthur2007k} algorithm, which can quickly converge by pre-setting initial cluster nodes. Additionally, both the OpenAI API and Claude API\footnote{
https://www.anthropic.com/api
} interfaces we use have an initial temperature setting of 0.3 for inference and planning. We choose $k = 1800$ in our experiment if no specific mention of its setting.

\begin{table}[]
\caption{Comparision of Win Rate of DFSDT, ToolChain* and Tool-Planner.}
\resizebox{1.0\textwidth}{!}{
\begin{tabular}{@{}ccccccccc@{}}
\toprule
Model  & Method       & Home Search & G1-Inst & G1-Tool & G1-Cat & G2-Inst & G2-Cat & G3-Inst \\ \midrule
GPT3.5 & DFSDT        & 71.0        & 58.3    & 59.8    & 56.8   & 70.5    & 59.3   & 68.0    \\
GPT3.5 & ToolChain*   & 69.0        & 56.3    & 58.5    & 53.0   & 67.5    & 58.0   & 61.5    \\
GPT3.5 & Tool-Planner & 75.0        & 63.3    & 63.5    & 61.3   & 72.3    & 63.5   & 67.5    \\
GPT4   & DFSDT        & 73.0        & 65.8    & 69.3    & 65.3   & 72.0    & 56.8   & 81.5    \\
GPT4   & ToolChain*   & 76.0        & 38.5    & 70.5    & 64.0   & 73.3    & 54.0   & 75.5    \\
GPT4   & Tool-Planner & 79.0        & 75.5    & 75.8    & 71.8   & 79.8    & 70.3   & 92.0    \\ \bottomrule
\end{tabular}
\label{tab:tc}
}
\end{table}

\section{Planning Exploration Details}

\label{title:C}

When planning and exploring a task, we call upon the existing tools based on the current plan and the information contained in the toolkit. The overall process can be illustrated in the form of the Algorithm~\ref{alg:tool}.

We utilize the lowest common ancestor on the decision tree to find the branching node that represents the common prefix toolkit of the two schemes, thereby achieving search complexity similar to DFS. In the specific implementation, we rely on the prompting for path reselection and search to expand the solution space.

\section{Comparison with Toolchain*}

ToolChain* rely on A* algorithms and require analysis through multiple tool invocations during path exploration. They are primarily applied to relatively small categories with a limited number of tools (only 10–100). Therefore, ToolChain* performs well in such scenarios. However, when the number of tools increases, such as in real-world scenarios or comprehensive benchmarks like APIBench and ToolBench, the performance of ToolChain* significantly degrades and may even become ineffective. Thus, our primary focus has been on detailed performance comparisons with strategy learning or planning-based methods. For a fair comparison, we have still reported ToolChain*'s performance on 1) the Home Search subtask and 2) the overall dataset. The comparison results are shown in the Table~\ref{tab:tc}. As the number of tools increases, the pass rate of heuristic-based methods declines significantly due to the massive reasoning required, whereas the Tool-Planner approach maintains competitive performance in terms of both win rate and pass rate in the final results.

\section{Tool-Planner on Smaller LLMs}

To understand the performance of Tool-Planner across different models, we explored the performance and effectiveness of Llama-2-13B within our framework. Since tool learning requires strong reasoning capabilities to understand the functionality of tool APIs and their documentation, DFSDT performs poorly with Llama-2-13B due to its inadequate comprehension of functionalities, rendering it incapable of effectively completing recent actions and generating effective plans. Additionally, during the generation process, Llama-2-13B is prone to hallucination issues \cite{guerreiro2023hallucinations, ahmad2023hallucinations} due to deficiencies in parametric knowledge. Our method integrates multiple APIs into a toolset, where the APIs within the toolset have the same or similar functions. This means we can use the functional description of the toolset to aid in reasoning and planning throughout the process. Notably, in Chain-of-Thought \cite{wei2022chain, liu2024era, fu2023complexitybased} scenarios, even small models exhibited excellent reasoning and planning capabilities. Therefore, it is beneficial to separately explore the impact and role of small models on reasoning and execution in this context. The experimental results are shown in the Table~\ref{tab:smallmodel}.

As we can see, DFSDT method has a success rate and win rate of zero due to its inadequate understanding of API documentation, resulting in generated content that cannot solve the problem. In contrast, within our framework using the Llama-2-13B model, it can generate some complete reasoning results, but it still fails in most cases with low generation quality. However, when we use the Llama-2-13B model as a planning model, the overall performance is not significantly different from using LLMs for reasoning. This indicates that the bottleneck for smaller LLMs in tool learning is mainly their ability to understand tools and their documentation, whereas they can achieve reasonably good planning with coarse-grained information, aiding in task planning for overall tool learning. Future work can delve deeper into this aspect.

\begin{table}[]
\caption{Evaluation of Pass Rate and Win Rate for planning models and behavior models of different sizes.}
\resizebox{1.0\textwidth}{!}{
\begin{tabular}{cccllllllllllllll}
\toprule
\multirow{2}{*}{\textbf{Method}} & \multirow{2}{*}{\textbf{Planning Model}} & \multirow{2}{*}{\textbf{Behavior Model}} & \multicolumn{2}{c}{\textbf{G1-Inst.}}                                  & \multicolumn{2}{c}{\textbf{G1-Tool.}}                                  & \multicolumn{2}{c}{\textbf{G1-Cat.}}                                   & \multicolumn{2}{c}{\textbf{G2-Inst.}}                                  & \multicolumn{2}{c}{\textbf{G2-Cat.}}                                   & \multicolumn{2}{c}{\textbf{G3-Inst.}}                                  & \multicolumn{2}{c}{\textbf{Avg.}}                                      \\ \cline{4-17} 
                                 &                                          &                                          & \multicolumn{1}{c}{\textbf{Pass.}} & \multicolumn{1}{c}{\textbf{Win.}} & \multicolumn{1}{c}{\textbf{Pass.}} & \multicolumn{1}{c}{\textbf{Win.}} & \multicolumn{1}{c}{\textbf{Pass.}} & \multicolumn{1}{c}{\textbf{Win.}} & \multicolumn{1}{c}{\textbf{Pass.}} & \multicolumn{1}{c}{\textbf{Win.}} & \multicolumn{1}{c}{\textbf{Pass.}} & \multicolumn{1}{c}{\textbf{Win.}} & \multicolumn{1}{c}{\textbf{Pass.}} & \multicolumn{1}{c}{\textbf{Win.}} & \multicolumn{1}{c}{\textbf{Pass.}} & \multicolumn{1}{c}{\textbf{Win.}} \\ \midrule
DFSDT                            & Llama-2-13B                              & Llama-2-13B                              & 0                                  & 0                                 & 0                                  & 0                                 & 0                                  & 0                                 & 0                                  & 0                                 & 0                                  & 0                                 & 0                                  & 0                                 & 0                                  & 0                                 \\
Tool-Planner                     & Llama-2-13B                              & Llama-2-13B                              & 12.5                               & 13.3                              & 8.5                                & 12.0                              & 9.5                                & 5.8                               & 11.5                               & 14.5                              & 9.5                                & 13.3                              & 14.0                               & 3.0                               & 10.9                               & 10.3                              \\
Tool-Planner                     & GPT-4                                    & Llama-2-13B                              & 14.5                               & 15.8                              & 7.0                                & 9.5                               & 10.5                               & 6.8                               & 9.5                                & 16.3                              & 7.0                                & 15.3                              & 16.0                               & 10.5                              & 10.8                               & 12.4                              \\
Tool-Planner                     & Llama-2-13B                              & GPT-4                                    & 62.5                               & 71.8                              & 74.5                               & 73.3                              & 71.5                               & 69.8                              & 80.0                               & 81.3                              & 71.5                               & 65.8                              & 79.0                               & 87.0                              & 73.1                               & 74.8                              \\
Tool-Planner                     & GPT-4                                    & GPT-4                                    & 66.0                               & 75.5                              & 78.5                               & 75.8                              & 75.0                               & 71.8                              & 83.5                               & 79.8                              & 77.5                               & 70.3                              & 83.0                               & 92.0                              & 77.3                               & 77.5                              \\ \bottomrule
\end{tabular}
\label{tab:smallmodel}
}
\end{table}

\section{Variant of Tool-Planner on DBSCAN Clustering}

Our approach replaces the original tool reasoning components with ReACT and AdaPlanner. First, we perform a preprocessing step involving tool clustering. Then, we allocate tools from the clustered toolkits to assist the model in completing sub-tasks within the planning process. This method is highly adaptable to scenarios that rely on policy-based learning and planning approaches.

The adaptive adjustment of the number of clusters allows us to fine-tune the clustering granularity within a specific range, such as by using the DBSCAN method. Our primary objective is to ensure that the overall reasoning process remains controllable, while predefining the number of clusters helps manage the time required for re-exploring the solution space. The advantage of DBSCAN lies in its ability to more accurately group tools with similar functionalities. Given that the encoded features contain more unsupervised semantic information in the vector directions, we use cosine similarity as the distance metric for DBSCAN and set the default $min_{pts}$ to 1.

To conduct a comprehensive comparison, as referenced in the Table~\ref{tab:cluster}, we evaluated the clustering performance of DBSCAN. The results indicate that DBSCAN achieved a slight performance improvement over methods like k-means++. Therefore, in scenarios where performance is the primary focus, DBSCAN performs better when the number of clusters is appropriate. Conversely, in scenarios that emphasize overall efficiency and consistency, the k-means++ method ensures stable performance and time consumption throughout the process.

\begin{table}[]
\caption{Performance and Efficiency Analysis of Clustering Methods in Tool-Planner Reasoning.}
\resizebox{1.0\textwidth}{!}{

\begin{tabular}{@{}cccccccc@{}}
\toprule
\textbf{Clustering Option}  & $\epsilon$ & G1-Inst & G2-Inst & G3-Inst & TorchHub & HuggingFace & TensorFlow \\ \midrule
Tool-Planner(k-means best) & -       & 63.3    & 72.3    & 67.5     & 77.6        & 75.4  &  72.4   \\
Tool-Planner(DBSCAN)       & 0.001   & 57.5    & 68.3    & 66.3     & 75.1        & 72.8  &  65.2   \\
Tool-Planner(DBSCAN)       & 0.01    & 61.8    & 71.8    & 66.5     & 77.3        & 76.5  &  70.3   \\
Tool-Planner(DBSCAN)       & 0.02    & \textbf{64.3}    & 70.3    & \textbf{68.0}     & 78.5        & 77.2  &  	71.9   \\
Tool-Planner(DBSCAN)       & 0.04    & 63.5    & 69.8    & 68.3     & 77.0        & 73.1  &   	69.7  \\
Tool-Planner(DBSCAN)       & 0.08    & 59.5    & 59.3    & 66.0     & 73.2        & 65.8  &   	61.2  \\
Tool-Planner(DBSCAN)       & 0.16    & 50.5    & 45.5    & 57.5     & 65.1        & 56.9  &  55.9   \\ \bottomrule
\end{tabular}
}
\label{tab:cluster}
\end{table}

\section{Future Exploration on Tools Effect with Causal Inference}

Considering that the success rate of tool invocation may vary under different tools, in multi-tool task planning, we aim to select the optimal combination of tools based on the current state and historical data to accomplish the task. We also expect to use causal inference models to predict the impact of different tool combinations on the task outcome, and in future work, evaluate whether these models contribute to optimizing tool selection.

 Let the probability of task success be \( P(Y = 1 \mid T_1, T_2, \dots, T_n) \), representing the probability of task success given the tool choices.
 This probability can be computed using a causal inference model, such as a regression model:
  \[
  P(Y = 1 \mid T_1, T_2, \dots, T_n) = \sigma\left( \sum_{i=1}^{n} \beta_i T_i + \sum_{i \neq j} \alpha_{ij} T_i T_j + \epsilon \right)
  \]
  where \( \beta_i \) and \( \alpha_{ij} \) represent the effects of individual tools and tool combinations, respectively, and \( \sigma(\cdot) \) is the Sigmoid function that outputs the probability of task success. If a tool combination significantly improves the task success probability, we can consider this tool combination as having greater potential for future tasks. Conversely, if certain tool combinations have little or even negative effects on task success, these combinations should be avoided in subsequent tasks. In the clustering distance evaluation, this design helps in selecting the correct tools and optimizing spatial classification. We look forward to further research in future work to improve clustering performance under k-means or DBSCAN, thereby reducing redundancy and misclassification in the tool selection process.

\section{Examples in Error Analysis}\label{error}

In this section, we give the examples for the issues directly leading to failure or having potential failure risks in each step to further conduct a more specific analysis.

\subsection{Invalid Input Prameters}

\begin{mybox}{Invalid Input Parameters: Example 1}
    \begin{itemize}
        \item \textbf{Example:} The user specifies a non-existent dietary preference when making a request to the RecipeAPI.
        \item \textbf{API Call:} 
        \begin{verbatim}
        {
            "taste_preferences": ["sweet", "spicy"],
            "dietary_preference": "paleo"
        }
        \end{verbatim}
        \item \textbf{Response:}
        \begin{lstlisting}
        {
            "error": "Invalid input parameters. Please provide a valid dietary preference."
        }
        \end{lstlisting}
        \item \textbf{Analysis:} In this scenario, the user has specified "paleo" as their dietary preference. However, the system does not recognize "paleo" as a valid dietary option in its predefined list of dietary preferences. As a result, the API call returns an error indicating that the input parameters are invalid.
    \end{itemize}
\end{mybox}

\begin{mybox}{Invalid Input Parameters: Example 2}
    \begin{itemize}
        \item \textbf{Example:} The user provides an invalid value for the `diet` parameter when making a recipe search request.
        \item \textbf{API Call:} 
        \begin{lstlisting}
        GET /recipes/search?cuisine=Italian&diet=glutenfull
        \end{lstlisting}
        \item \textbf{Response:}
        \begin{lstlisting}
        {
            "error": "Invalid input parameters. The value 'glutenfull' for 'diet' is not recognized. Please use a valid diet option such as 'glutenfree', 'vegetarian', or 'vegan'."
        }
        \end{lstlisting}
        \item \textbf{Analysis:} In this scenario, the user provided `glutenfull` as the value for the `diet` parameter. However, `glutenfull` is not a recognized or valid option in the predefined list of dietary preferences. Valid options might include `glutenfree`, `vegetarian`, or `vegan`. As a result, the API returns an error indicating that the input parameters are invalid.
    \end{itemize}
\end{mybox}
\subsection{API Hallucinated}
\begin{mybox}{API Hallucinated: Example 1}
    \begin{itemize}
        \item \textbf{Example:} The user incorrectly calls a deprecated API endpoint.
        \item \textbf{API Call:} 
        \begin{verbatim}
        POST /v1/recipes/search
        \end{verbatim}
        \item \textbf{Response:}
        \begin{lstlisting}
        {
            "error": "API endpoint not found. Please use the
            updated API endpoint /v2/recipes/search."
        }
        \end{lstlisting}
        \item \textbf{Analysis:} In this example, the user attempts to call a deprecated API endpoint `/v1/recipes/search`. However, the system has updated the API and uses `/v2/recipes/search` as a replacement. Therefore, the user receives an error response indicating they have used a non-existent API endpoint.
    \end{itemize}
\end{mybox}

\begin{mybox}{API Hallucinated: Example 2}
    \begin{itemize}
        \item \textbf{Example:} The model misinterprets information from the API documentation, leading to a call to a non-existent API.
        \item \textbf{API Call:} 
        \begin{verbatim}
        GET /recipes/retrieve?cuisine=Italian
        \end{verbatim}
        \item \textbf{Response:}
        \begin{lstlisting}
        {
            "error": "API endpoint not found. Please refer to 
            the correct documentation for available endpoints."
        }
        \end{lstlisting}
        \item \textbf{Analysis:} In this example, the user attempts to use a non-existent API endpoint `/recipes/retrieve` mentioned in the documentation to retrieve recipes for Italian cuisine. However, this endpoint does not exist. This could be because the model misinterprets information from the API documentation, leading the user to mistakenly call a non-existent API.
    \end{itemize}
\end{mybox}
\subsection{False API Call Format }
\begin{mybox}{False API Call Format: Example 1}
    \begin{itemize}
        \item \textbf{Example:} The user sends parameters in the request body instead of as query parameters for a GET request.
        \item \textbf{API Call:} 
        \begin{lstlisting}
        GET /recipes/search
        {
            "cuisine": "Italian",
            "diet": "vegetarian"
        }
        \end{lstlisting}
        \item \textbf{Response:}
        \begin{lstlisting}
        {
            "error": "Invalid request format. Please provide parameters as query strings."
        }
        \end{lstlisting}
        \item \textbf{Analysis:} In this case, the user included parameters in the request body for a GET request. The API expects parameters to be sent as query strings, leading to an invalid request format error.
    \end{itemize}
\end{mybox}

\begin{mybox}{False API Call Format: Example 2}
    \begin{itemize}
        \item \textbf{Example:} The user incorrectly formats the date parameter when making a request for recipe suggestions.
        \item \textbf{API Call:} 
        \begin{lstlisting}
        {
            "user_id": "12345",
            "preferred_date": "12-31-2023"
        }
        \end{lstlisting}
        \item \textbf{Response:}
        \begin{lstlisting}
        {
            "error": "Invalid input parameters. Please use the format YYYY-MM-DD for the date."
        }
        \end{lstlisting}
        \item \textbf{Analysis:} In this case, the user provided the date in the format MM-DD-YYYY instead of the expected format YYYY-MM-DD. This mismatch in expected format led to an invalid input parameters error.
    \end{itemize}
\end{mybox}
\subsection{Cluster Incomplete}
\textit{Cluster Incomplete} error occurs when the model overlooks some APIs due to incomplete clustering, resulting in the failure to identify APIs that could solve the problem. 
\subsection{Miss Input Parameters}
\begin{mybox}{Miss Input Parameters: Example 1}
    \begin{itemize}
        \item \textbf{Example:} The user omits the required `cuisine` parameter when making a recipe search request.
        \item \textbf{API Call:} 
        \begin{lstlisting}
        GET /recipes/search?diet=vegetarian
        \end{lstlisting}
        \item \textbf{Response:}
        \begin{lstlisting}
        {
            "error": "Missing required parameter: cuisine. Please provide a cuisine type."
        }
        \end{lstlisting}
        \item \textbf{Analysis:} The error occurred because the user did not include the required `cuisine` parameter in the query string. This parameter is essential for the API to filter and return the appropriate recipes, and its absence leads to an error response.
    \end{itemize}
\end{mybox}

\begin{mybox}{Miss Input Parameters: Example 2}
    \begin{itemize}
        \item \textbf{Example:} The user fails to provide the `user\_id` when requesting personalized recipe recommendations.
        \item \textbf{API Call:} 
        \begin{lstlisting}
        POST /recipes/recommendations
        {
            "preferences": ["spicy", "low-carb"]
        }
        \end{lstlisting}
        \item \textbf{Response:}
        \begin{lstlisting}
        {
            "error": "Missing required parameter: user_id. Please provide your user ID."
        }
        \end{lstlisting}
        \item \textbf{Analysis:} In this case, the user did not include the `user\_id` in the request body. The `user\_id` is necessary for the API to retrieve personalized recommendations based on the user's history and preferences. Without it, the API cannot process the request and returns an error.
    \end{itemize}
\end{mybox}
\subsection{Decision Failure}
The "Decision Failure" error refers to situations where, even if the aforementioned errors don't happen, failure still occurs. We attribute these errors to flaws in the model's planning or decision-making, which prevent the completion of the intended task. This could be because the model doesn't fully understand the task or the user's context, leading to a lack of necessary steps in the planned process or the occurrence of illogical errors. This error requires strengthening the model's understanding to avoid happening.

\section{Ethics and Safeguard}

Our work is based on open-source datasets and code for experimentation. All data and information comply with relevant code standards and data regulations, ensuring that there is no risk of privacy breaches or information leaks. The use of large language models in our paper is primarily applied to handling and solving task planning processes in most scenarios, as well as text processing during Chinese-to-English translation in token level. This fully complies with the conference's requirements and privacy security guidelines.

When using tools and interacting with large language models, we may utilize relevant information from instruction. It is important to note that hallucinations from large language models may lead to incorrect answers. Our approach can be further integrated into other frameworks.

\section{Tool-Planner Prompting Template}
\begin{tcolorbox}[colback=orange!5!white,colframe=orange!50!black,title=Prompt of Plan Making]
You will be provided with the toolkits, the clustered names of toolkits, and the descriptions of the function of the toolkits.Your task is to interact with API toolkits to construct user queries and use the functionalities of the toolkits to answer the queries. You need to identify the most suitable toolkits based on the user's requirements, and then outline your solution plan based on the toolkits you've selected.Remember, your goal is not to directly answer the query but to identify the toolkits and provide a solution plan. Here is the user's question:\textbf{[user query]}
\end{tcolorbox}
\begin{tcolorbox}[colback=orange!5!white,colframe=orange!50!black,title=Prompt of Plan Exploration]
Let's begin executing this step of the plan. You will be provided with documentation for all the APIs contained within this step's toolkit, along with the parameters required to call the APIs. Please randomly select one API from this toolkit to satisfy the user's requirements, or select the specified API if the user has indicated one. Consult the usage documentation for this API, then make the API call and provide the response. Afterward, briefly analyze the current status and determine the next step. If the API call is successful, proceed to the next step as planned. If it fails, invoke another API from the toolkit. If all APIs in the toolkit have been tried and failed, revert to the previous node and revise this step. Keep the analysis concise, ideally no more than three sentences.
\end{tcolorbox}

\begin{tcolorbox}[colback=orange!5!white,colframe=orange!50!black,title=Prompt of the In-Toolkit Error Occurs]
This is not your first attempt at this task. The previously called APIs have all failed, and you are now in the intermediate state of an In-Toolkit plan exploration. Before you decide on new actions, I will first show you the actions you have taken previously for this state. Then, you must develop an action that is different from all these previous actions. Here are some previous candidate actions: [previous API]. Now, please analyze the current state and then call another API within the same toolkit where the previously failed APIs are located.
\end{tcolorbox}
\begin{tcolorbox}[colback=orange!5!white,colframe=orange!50!black,title=Prompt of the Cross-Toolkit Error Occurs]
This is not your first attempt at this task. All the APIs planned within the previous toolkits have failed, and you are now in the intermediate state of a Cross-Toolkit plan exploration. Before you decide on new actions, I will first show you the actions you have taken previously for this state. Then, you must develop an action that is different from all these previous actions. Here are some previous candidate actions: [previous API, previous toolkit]. Now, please revert to the previous node, revise the plan for this step, and use a different toolkit.
\end{tcolorbox}
\begin{tcolorbox}[colback=orange!5!white,colframe=orange!50!black,title=Prompt of the Ouputs]
If you believe you have obtained the result capable of answering the task, please invoke this function to provide the final answer. Remember: the only part displayed to the user is the final answer, so it should contain sufficient information.
\end{tcolorbox}
\section{Case Study}
\subsection{An example in G1 datasets} 
\subsubsection{Plan making}
\begin{instructionbox}
You will be provided with the toolkits, the clustered names of toolkits, and the descriptions of the function of the toolkits.Your task is to interact with API toolkits to construct user queries and use the functionalities of the toolkits to answer the queries. You need to identify the most suitable toolkits based on the user's requirements, and then outline your solution plan based on the toolkits you've selected.Remember, your goal is not to directly answer the query but to identify the toolkits and provide a solution plan. Here is the user's question: I'm planning to move to a new city. Can you suggest some locations to consider based on my preferences? I prefer a quiet neighborhood with at least 3 bedrooms and 2 bathrooms. It would be great if the location is close to parks and has a low crime rate. Also, provide me with the zip codes for the suggested locations using the getZipCodeByCity API.
\end{instructionbox}

\begin{answerbox}
\begin{enumerate}
    \item \textbf{Neighborhood Characteristics and Listings}:
        \begin{itemize}
            \item \textbf{Proposed Toolkit Name:} Real Estate Listings
                \begin{itemize}
                    \item \textbf{Functionality:} The APIs in the Toolkit help us find listings with at least 3 bedrooms and 2 bathrooms in quiet neighborhoods.
                \end{itemize}
        \end{itemize}
        
    \item \textbf{Crime Rate Information}:
        \begin{itemize}
            \item \textbf{Proposed Toolkit Name:} Crime Data
                \begin{itemize}
                    \item \textbf{Functionality:} The APIs in the Toolkit provide information on the crime rates of different neighborhoods.
                \end{itemize}
        \end{itemize}
        
    \item \textbf{Proximity to Parks}:
        \begin{itemize}
            \item \textbf{Proposed Toolkit Name:} Local Amenities
                \begin{itemize}
                    \item \textbf{Functionality:} The APIs in the Toolkit provide data on the location of parks and recreational facilities.
                \end{itemize}
        \end{itemize}
        
    \item \textbf{Zip Code Information}:
        \begin{itemize}
            \item \textbf{Proposed Toolkit Name:} Geolocation
                \begin{itemize}
                    \item \textbf{Functionality:} The APIs in the Toolkit return the zip codes for the suggested neighborhoods.
                \end{itemize}
        \end{itemize}
\end{enumerate}
\end{answerbox}
\subsubsection{Plan Exploration}
\begin{instructionbox}
Let's begin executing this step of the plan. You will be provided with documentation for all the APIs contained within this step's toolkit, along with the parameters required to call the APIs. Please randomly select one API from this toolkit to satisfy the user's requirements, or select the specified API if the user has indicated one. Consult the usage documentation for this API, then make the API call and provide the response. Afterward, briefly analyze the current status and determine the next step. If the API call is successful, proceed to the next step as planned. If it fails, invoke another API from the toolkit. If all APIs in the toolkit have been tried and failed, revert to the previous node and revise this step. Keep the analysis concise, ideally no more than three sentences.
\end{instructionbox}

\begin{answerbox}
\textbf{Step 1: Identify Suitable Neighborhoods}

To identify suitable neighborhoods, we will use the \textbf{Property Listings API} to search for listings with at least 3 bedrooms and 2 bathrooms in quiet neighborhoods. Based on our toolkit, here is the likely API call:

\textbf{API Call: Property Listings API}
\begin{verbatim}
{
  "endpoint": "/getProperties",
  "method": "GET",
  "parameters": {
    "bedrooms": 3,
    "bathrooms": 2,
    "neighborhood": "quiet"
  }
}
\end{verbatim}

\textbf{API Response:}
\begin{verbatim}
{
  "status": "success",
  "data": [
    {
      "address": "123 Quiet St, Springfield",
      "city": "Springfield",
      "bedrooms": 3,
      "bathrooms": 2,
      "neighborhood": "Quiet Park",
      "price": 350000
    },
    {
      "address": "456 Peaceful Ave, Springfield",
      "city": "Springfield",
      "bedrooms": 4,
      "bathrooms": 3,
      "neighborhood": "Serenity Gardens",
      "price": 450000
    }
  ]
}
\end{verbatim}

\textbf{Analysis:}
The API call was successful and returned a list of properties matching the criteria. We now have data on potential neighborhoods and properties.

\textbf{Next Step:}
Proceed to Step 2 of the plan: verifying the crime rates in the identified neighborhoods.
\end{answerbox}

\begin{instructionbox}
\textbf{Instruction}: Let's begin executing this step of the plan. You will be provided with documentation for all the APIs contained within this step's toolkit, along with the parameters required to call the APIs. Please randomly select one API from this toolkit to satisfy the user's requirements, or select the specified API if the user has indicated one. Consult the usage documentation for this API, then make the API call and provide the response. Afterward, briefly analyze the current status and determine the next step. If the API call is successful, proceed to the next step as planned. If it fails, invoke another API from the toolkit. If all APIs in the toolkit have been tried and failed, revert to the previous node and revise this step. Keep the analysis concise, ideally no more than three sentences.
\end{instructionbox}

\begin{answerbox}
\textbf{{Step 2: Verify Crime Rates}}

We will use the \textbf{Crime Rate API} to check the crime rates for the neighborhoods "Quiet Park" and "Serenity Gardens".

\textbf{API Call: Crime Rate API}
\begin{verbatim}
{
  "endpoint": "/getCrimeRate",
  "method": "GET",
  "parameters": {
    "neighborhood": "Quiet Park"
  }
}
\end{verbatim}

\textbf{{API Responses:}}
\begin{verbatim}
{
  "status": "success",
  "data": {
    "Quiet Park": {
      "crime_rate": "low"
    },
    "Serenity Gardens": {
      "crime_rate": "low"
    }
  }
}
\end{verbatim}

\textbf{{Analysis:}}
The crime rate data indicates that both "Quiet Park" and "Serenity Gardens" have low crime rates, making them suitable neighborhoods based on your preferences.

\textbf{{Next Step:}}
Proceed to Step 3 of the plan: finding the proximity to parks in these neighborhoods.
\end{answerbox}

\begin{instructionbox}
\textbf{Instruction}: Let's begin executing this step of the plan. You will be provided with documentation for all the APIs contained within this step's toolkit, along with the parameters required to call the APIs. Please randomly select one API from this toolkit to satisfy the user's requirements, or select the specified API if the user has indicated one. Consult the usage documentation for this API, then make the API call and provide the response. Afterward, briefly analyze the current status and determine the next step. If the API call is successful, proceed to the next step as planned. If it fails, invoke another API from the toolkit. If all APIs in the toolkit have been tried and failed, revert to the previous node and revise this step. Keep the analysis concise, ideally no more than three sentences.
\end{instructionbox}

\begin{answerbox}
\textbf{Step 3: Find Proximity to Parks}

To find the proximity of parks in the neighborhoods "Quiet Park" and "Serenity Gardens", we will use the \textbf{Parks and Recreation API}.

\textbf{API Call: Parks and Recreation API}
\begin{verbatim}
{
  "endpoint": "/getParksNearby",
  "method": "GET",
  "parameters": {
    "neighborhood": "Quiet Park"
  }
}
\end{verbatim}

\textbf{API Response:}
\begin{verbatim}
{
  "status": "success",
  "data": {
    "Quiet Park": [
      {
        "park_name": "Central Park",
        "distance": "0.5 miles"
      },
      {
        "park_name": "Riverside Park",
        "distance": "1.2 miles"
      }
    ],
    "Serenity Gardens": [
      {
        "park_name": "Greenfield Park",
        "distance": "0.3 miles"
      },
      {
        "park_name": "Lakeside Park",
        "distance": "1.0 miles"
      }
    ]
  }
}
\end{verbatim}

\textbf{Analysis:}
Both "Quiet Park" and "Serenity Gardens" neighborhoods have parks in close proximity, further meeting your criteria for a suitable location.

\textbf{Next Step:}
Proceed to Step 4 of the plan: retrieve the zip codes for these neighborhoods using the \textbf{getZipCodeByCity API}.
\end{answerbox}

\begin{instructionbox}
\textbf{Instruction}: Let's begin executing this step of the plan. You will be provided with documentation for all the APIs contained within this step's toolkit, along with the parameters required to call the APIs. Please randomly select one API from this toolkit to satisfy the user's requirements, or select the specified API if the user has indicated one. Consult the usage documentation for this API, then make the API call and provide the response. Afterward, briefly analyze the current status and determine the next step. If the API call is successful, proceed to the next step as planned. If it fails, invoke another API from the toolkit. If all APIs in the toolkit have been tried and failed, revert to the previous node and revise this step. Keep the analysis concise, ideally no more than three sentences.
\end{instructionbox}

\begin{answerbox}
\textbf{Step 4: Retrieve Zip Codes}

We will use the \textbf{getZipCodeByCity API} to get the zip codes for "Quiet Park" and "Serenity Gardens".

\textbf{API Call: getZipCodeByCity API}
\begin{verbatim}
{
  "endpoint": "/getZipCode",
  "method": "GET",
  "parameters": {
    "city": "Springfield",
    "neighborhood": "Quiet Park"
  }
}
\end{verbatim}

\textbf{API Response:}
\begin{verbatim}
{
  "status": "success",
  "data": {
    "Quiet Park": "62704",
    "Serenity Gardens": "62711"
  }
}
\end{verbatim}

\textbf{Analysis:}
The zip codes for the neighborhoods "Quiet Park" (62704) and "Serenity Gardens" (62711) have been successfully retrieved.
\end{answerbox}

\subsubsection{Outputs}
\begin{instructionbox}
    If you believe you have obtained the result capable of answering the task, please invoke this function to provide the final answer. Remember: the only part displayed to the user is the final answer, so it should contain sufficient information.
\end{instructionbox}

\begin{outputbox}
Based on your preferences for a quiet neighborhood with at least 3 bedrooms and 2 bathrooms, proximity to parks, and a low crime rate, the neighborhoods "Quiet Park" and "Serenity Gardens" in Springfield are suitable options. The respective zip codes are 62704 and 62711.
\end{outputbox}

\subsection{An example in G2 Datasets}
\subsubsection{Plan Making}
\begin{instructionbox}
You will be provided with the toolkits, the clustered names of toolkits, and the descriptions of the function of the toolkits.Your task is to interact with API toolkits to construct user queries and use the functionalities of the toolkits to answer the queries. You need to identify the most suitable toolkits based on the user's requirements, and then outline your solution plan based on the toolkits you've selected.Remember, your goal is not to directly answer the query but to identify the toolkits and provide a solution plan. Here is the user's question:I want to surprise my sister with a trip to New York City. Can you find the average gas price in the city? Additionally, fetch the helicopter specifications for the Bell 206 model as my sister is fascinated by helicopters. Also, provide me with the gas price predictions for tomorrow in major Canadian cities.
\end{instructionbox}
\begin{answerbox}
\begin{enumerate}
    \item \textbf{Average Gas Price in New York City}:
        \begin{itemize}
            \item \textbf{Proposed Toolkit Name:} Gas Prices
                \begin{itemize}
                    \item \textbf{Functionality:} The APIs in the Toolkit will help us find the current average gas price in New York City.
                \end{itemize}
        \end{itemize}

    \item \textbf{Helicopter Specifications for Bell 206}:
        \begin{itemize}
            \item \textbf{Proposed Toolkit Name:} Helicopter Specifications
                \begin{itemize}
                    \item \textbf{Functionality:} The APIs in the Toolkit will provide detailed specifications for the Bell 206 helicopter model.
                \end{itemize}
        \end{itemize}
        
    \item \textbf{Gas Price Predictions in Major Canadian Cities}:
        \begin{itemize}
            \item \textbf{Proposed Toolkit Name:} Gas Price Predictions
                \begin{itemize}
                    \item \textbf{Functionality:} The APIs in the Toolkit will provide predictions for gas prices for the next day in major Canadian cities.
                \end{itemize}
        \end{itemize}
\end{enumerate}
\end{answerbox}

\subsubsection{Plan Exploration}
\begin{instructionbox}
Let's begin executing this step of the plan. You will be provided with documentation for all the APIs contained within this step's toolkit, along with the parameters required to call the APIs. Please randomly select one API from this toolkit to satisfy the user's requirements, or select the specified API if the user has indicated one. Consult the usage documentation for this API, then make the API call and provide the response. Afterward, briefly analyze the current status and determine the next step. If the API call is successful, proceed to the next step as planned. If it fails, invoke another API from the toolkit. If all APIs in the toolkit have been tried and failed, revert to the previous node and revise this step. Keep the analysis concise, ideally no more than three sentences.
\end{instructionbox}
\begin{answerbox}

\textbf{Step 1: Get Average Gas Price in New York City}

To find the average gas price in New York City, we will use the \textbf{GasPriceAPI}. Based on our toolkit, here is the likely API call:

\textbf{API Call: GasPriceAPI}
\begin{verbatim}
{
  "endpoint": "/getGasPrice",
  "method": "GET",
  "parameters": {
    "cityName": "New York City"
  }
}
\end{verbatim}

\textbf{API Response:}
\begin{verbatim}
{
  "status": "success",
  "data": {
    "city": "New York City",
    "averageGasPrice": "$3.50 per gallon"
  }
}
\end{verbatim}

\textbf{Analysis:}
The API call was successful and returned the average gas price in New York City as \$3.50 per gallon. 

\textbf{Next Step:}
Proceed to Step 2 of the plan: fetching the helicopter specifications for the Bell 206 model.
\end{answerbox}

\begin{instructionbox}
Let's begin executing this step of the plan. You will be provided with documentation for all the APIs contained within this step's toolkit, along with the parameters required to call the APIs. Please randomly select one API from this toolkit to satisfy the user's requirements, or select the specified API if the user has indicated one. Consult the usage documentation for this API, then make the API call and provide the response. Afterward, briefly analyze the current status and determine the next step. If the API call is successful, proceed to the next step as planned. If it fails, invoke another API from the toolkit. If all APIs in the toolkit have been tried and failed, revert to the previous node and revise this step. Keep the analysis concise, ideally no more than three sentences.
\end{instructionbox}
\begin{answerbox}
\textbf{Step 2: Fetch Helicopter Specifications for Bell 206}

To fetch the helicopter specifications for the Bell 206 model, we will use the \textbf{HelicopterSpecsAPI}. Based on our toolkit, here is the likely API call:

\textbf{API Call: HelicopterSpecsAPI}
\begin{verbatim}
{
  "endpoint": "/getHelicopterSpecs",
  "method": "GET",
  "parameters": {
    "model": "Bell 206"
  }
}
\end{verbatim}

\textbf{API Response:}
\begin{verbatim}
{
  "status": "success",
  "data": {
    "model": "Bell 206",
    "specifications": {
      "length": "39 ft 8 in",
      "height": "9 ft 4 in",
      "maxSpeed": "150 mph",
      "range": "430 miles"
    }
  }
}
\end{verbatim}

\textbf{Analysis:}
The API call was successful and returned the specifications for the Bell 206 helicopter. 

\textbf{Next Step:}
Proceed to Step 3 of the plan: getting gas price predictions for tomorrow in major Canadian cities.
\end{answerbox}
\begin{instructionbox}
Let's begin executing this step of the plan. You will be provided with documentation for all the APIs contained within this step's toolkit, along with the parameters required to call the APIs. Please randomly select one API from this toolkit to satisfy the user's requirements, or select the specified API if the user has indicated one. Consult the usage documentation for this API, then make the API call and provide the response. Afterward, briefly analyze the current status and determine the next step. If the API call is successful, proceed to the next step as planned. If it fails, invoke another API from the toolkit. If all APIs in the toolkit have been tried and failed, revert to the previous node and revise this step. Keep the analysis concise, ideally no more than three sentences.
\end{instructionbox}
\begin{answerbox}

\textbf{Step 3: Get Gas Price Predictions for Tomorrow in Major Canadian Cities}

To get gas price predictions for tomorrow in major Canadian cities, we will use the \textbf{GasPricePredictionAPI}. Based on our toolkit, here is the likely API call:

\textbf{API Call: GasPricePredictionAPI}
\begin{verbatim}
{
  "endpoint": "/getGasPricePredictions",
  "method": "GET",
  "parameters": {
    "country": "Canada"
  }
}
\end{verbatim}

\textbf{API Response:}
\begin{verbatim}
{
  "status": "success",
  "data": {
    "country": "Canada",
    "predictions": [
      {
        "city": "Toronto",
        "predictedPrice": "$1.40 per liter"
      },
      {
        "city": "Vancouver",
        "predictedPrice": "$1.45 per liter"
      }
    ]
  }
}
\end{verbatim}

\textbf{Analysis:}
The API call was successful and returned gas price predictions for major Canadian cities including Toronto and Vancouver.

\textbf{Next Step:}
Present the retrieved data to the user.
\end{answerbox}
\subsubsection{Outputs}
\begin{instructionbox}
    If you believe you have obtained the result capable of answering the task, please invoke this function to provide the final answer. Remember: the only part displayed to the user is the final answer, so it should contain sufficient information.
\end{instructionbox}
\begin{outputbox}
    \begin{itemize}
\item \textbf{Average Gas Price in New York City}:
\begin{itemize}
\item \textbf{City:} New York City
\item \textbf{Average Gas Price:} \$3.50 per gallon
\end{itemize}
\item \textbf{Helicopter Specifications for Bell 206}:
\begin{itemize}
\item \textbf{Model:} Bell 206
\item \textbf{Specifications:}
\begin{itemize}
\item \textbf{Length:} 39 ft 8 in
\item \textbf{Height:} 9 ft 4 in
\item \textbf{Max Speed:} 150 mph
\item \textbf{Range:} 430 miles
\end{itemize}
\end{itemize}
\item \textbf{Gas Price Predictions in Major Canadian Cities}:
\begin{itemize}
\item \textbf{Country:} Canada
\item \textbf{Predictions:}
\begin{itemize}
\item \textbf{Toronto:} \$1.40 per liter
\item \textbf{Vancouver:} \$1.45 per liter
\end{itemize}
\end{itemize}
\end{itemize}
\end{outputbox}

\subsection{An example in G3 Datasets}
\subsubsection{Plan Making}
\begin{instructionbox}
You will be provided with the toolkits, the clustered names of toolkits, and the descriptions of the function of the toolkits.Your task is to interact with API toolkits to construct user queries and use the functionalities of the toolkits to answer the queries. You need to identify the most suitable toolkits based on the user's requirements, and then outline your solution plan based on the toolkits you've selected.Remember, your goal is not to directly answer the query but to identify the toolkits and provide a solution plan. Here is the user's question: I want to surprise my sister with a trip to New York City. Can you find the average gas price in the city? Additionally, fetch the helicopter specifications for the Bell 206 model as my sister is fascinated by helicopters. Also, provide me with the gas price predictions for tomorrow in major Canadian cities.
\end{instructionbox}
\begin{answerbox}
\begin{enumerate}
    \item \textbf{Beach Destination Information}:
        \begin{itemize}
            \item \textbf{Proposed Toolkit Name:} Travel Information
                \begin{itemize}
                    \item \textbf{Functionality:} The APIs in the Toolkit provide information on popular beach destinations, including weather forecasts and local attractions.
                \end{itemize}
        \end{itemize}

    \item \textbf{Cocktail Recommendations}:
        \begin{itemize}
            \item \textbf{Proposed Toolkit Name:} Recipes
                \begin{itemize}
                    \item \textbf{Functionality:} The APIs in the Toolkit offer a database of cocktail recipes, including refreshing options suitable for beach enjoyment.
                \end{itemize}
        \end{itemize}

    \item \textbf{Stock Price Data}:
        \begin{itemize}
            \item \textbf{Proposed Toolkit Name:} Financial Markets
                \begin{itemize}
                    \item \textbf{Functionality:} The APIs in the Toolkit provide historical open and close prices for specific stocks, including Qualcomm.
                \end{itemize}
        \end{itemize}

    \item \textbf{Beach Activities Suggestions}:
        \begin{itemize}
            \item \textbf{Proposed Toolkit Name:} Recreational Activities
                \begin{itemize}
                    \item \textbf{Functionality:} The APIs in the Toolkit offer information on various sports and recreational activities suitable for beach settings.
                \end{itemize}
        \end{itemize}
\end{enumerate}
\end{answerbox}

\subsubsection{Plan Exploration}
\begin{instructionbox}
Let's begin executing this step of the plan. You will be provided with documentation for all the APIs contained within this step's toolkit, along with the parameters required to call the APIs. Please randomly select one API from this toolkit to satisfy the user's requirements, or select the specified API if the user has indicated one. Consult the usage documentation for this API, then make the API call and provide the response. Afterward, briefly analyze the current status and determine the next step. If the API call is successful, proceed to the next step as planned. If it fails, invoke another API from the toolkit. If all APIs in the toolkit have been tried and failed, revert to the previous node and revise this step. Keep the analysis concise, ideally no more than three sentences.
\end{instructionbox}

\begin{answerbox}
\textbf{Step 1: Gather Beach Destination Information}

To gather information on popular beach destinations, including weather forecasts and local attractions, we will use the \textbf{Travel Advisory API}. Here is the likely API call:

\textbf{API Call: Travel Advisory API}
\begin{verbatim}
{
  "endpoint": "/getBeachDestinations",
  "method": "GET",
  "parameters": {
    "location": "beach"
  }
}
\end{verbatim}

\textbf{API Response:}
\begin{lstlisting}[breaklines=true]
{
  "status": "success",
  "data": [
    {
      "destination": "Maldives",
      "weather_forecast": "Sunny with occasional showers, temperature: 30 degree",
      "attractions": ["Snorkeling", "Scuba Diving", "Relaxing on Beaches"]
    },
    {
      "destination": "Maui, Hawaii",
      "weather_forecast": "Partly cloudy, temperature: 28 degree",
      "attractions": ["Hiking", "Whale Watching", "Luaus"]
    },
    // More beach destinations...
  ]
}
\end{lstlisting}

\textbf{Analysis:}
The API call successfully retrieved information on popular beach destinations, including weather forecasts and local attractions, such as the Maldives and Maui, Hawaii.

\textbf{Next Step:}
Proceed to Step 2 of the plan: recommending refreshing cocktails suitable for beach enjoyment.
\end{answerbox}

\begin{instructionbox}
Let's begin executing this step of the plan. You will be provided with documentation for all the APIs contained within this step's toolkit, along with the parameters required to call the APIs. Please randomly select one API from this toolkit to satisfy the user's requirements, or select the specified API if the user has indicated one. Consult the usage documentation for this API, then make the API call and provide the response. Afterward, briefly analyze the current status and determine the next step. If the API call is successful, proceed to the next step as planned. If it fails, invoke another API from the toolkit. If all APIs in the toolkit have been tried and failed, revert to the previous node and revise this step. Keep the analysis concise, ideally no more than three sentences.
\end{instructionbox}
\begin{answerbox}

\textbf{Step 2: Recommend Refreshing Cocktails}

To suggest refreshing cocktails suitable for beach enjoyment, we will use the \textbf{Cocktail Recipe API}. Here is the likely API call:

\textbf{API Call: Cocktail Recipe API}
\begin{verbatim}
{
"endpoint": "/getRefreshingCocktails",
"method": "GET",
"parameters": {
"location": "beach"
}
}
\end{verbatim}

\textbf{API Response:}
\begin{verbatim}
{
"status": "error",
"message": "Internal server error. Please try again later."
}
\end{verbatim}

\textbf{Analysis:}
The API call to the original Cocktail Recipe API encountered an internal server error, resulting in an unsuccessful response. This could be due to temporary server issues or an invalid request format.

\textbf{Next Step:}
To address this issue, we'll switch to another API within our toolkit. Let's replace the current API with the CocktailDB API, which provides a broader selection of cocktail recipes.
\end{answerbox}

\begin{instructionbox}
    This is not your first attempt at this task. The previously called APIs have all failed, and you are now in the intermediate state of an In-Toolkit plan exploration. Before you decide on new actions, I will first show you the actions you have taken previously for this state. Then, you must develop an action that is different from all these previous actions. Here are some previous candidate actions: [Cocktail Recipe API]. Now, please analyze the current state and then call another API within the same toolkit where the previously failed APIs are located.
\end{instructionbox}
\begin{answerbox}
\textbf{Step 2 (Revised): Recommend Refreshing Cocktails}

We'll use the \textbf{CocktailDB API} instead of the problematic API. Here is the likely API call:

\textbf{API Call: CocktailDB API}
\begin{verbatim}
{
"endpoint": "/getRandomCocktails",
"method": "GET",
"parameters": {
"number": 5,
"tag": "beach"
}
}
\end{verbatim}

\textbf{API Response:}
\begin{lstlisting}[breaklines=true]
{
"status": "success",
"data": [
{
"cocktail_name": "Sex on the Beach",
"ingredients": ["Vodka", "Peach Schnapps", "Orange Juice", "Cranberry Juice"],
"instructions": "Pour all ingredients into a shaker with ice. Shake well and strain into a glass."
},
{
"cocktail_name": "Tequila Sunrise",
"ingredients": ["Tequila", "Orange Juice", "Grenadine"],
"instructions": "Pour tequila and orange juice into a glass over ice. Add grenadine and stir."
},
// More refreshing cocktails...
]
}
\end{lstlisting}

\textbf{Analysis:}
The API call successfully retrieved a wider variety of refreshing cocktail recipes suitable for beach enjoyment, including classics like Sex on the Beach and Tequila Sunrise.

\textbf{Next Step:}
Proceed to Step 3 of the plan: fetching historical stock prices for Qualcomm.
\end{answerbox}
\begin{instructionbox}
Let's begin executing this step of the plan. You will be provided with documentation for all the APIs contained within this step's toolkit, along with the parameters required to call the APIs. Please randomly select one API from this toolkit to satisfy the user's requirements, or select the specified API if the user has indicated one. Consult the usage documentation for this API, then make the API call and provide the response. Afterward, briefly analyze the current status and determine the next step. If the API call is successful, proceed to the next step as planned. If it fails, invoke another API from the toolkit. If all APIs in the toolkit have been tried and failed, revert to the previous node and revise this step. Keep the analysis concise, ideally no more than three sentences.
\end{instructionbox}
\begin{answerbox}

\textbf{Step 3: Fetch Historical Stock Prices for Qualcomm}

To retrieve historical stock prices for Qualcomm, we'll utilize the \textbf{Alpha Vantage API}, which offers comprehensive financial market data. Here is the likely API call:

\textbf{API Call: Alpha Vantage API}
\begin{verbatim}
{
  "endpoint": "/time_series_daily",
  "method": "GET",
  "parameters": {
    "symbol": "QCOM",
    "outputsize": "compact"
  }
}
\end{verbatim}

\textbf{API Response:}
\begin{verbatim}
{
  "status": "success",
  "data": [
    {
      "date": "2024-05-20",
      "open": 155.00,
      "high": 156.20,
      "low": 153.50,
      "close": 154.75,
      "volume": 2500000
    },
    // More historical data...
  ]
}
\end{verbatim}

\textbf{Analysis:}
The API call successfully retrieved historical stock prices for Qualcomm, including open, high, low, and close prices for each trading day.

\textbf{Next Step:}
Proceed to Step 4 of the plan: suggest sports activities that can be enjoyed on the beach.
\end{answerbox}
\begin{instructionbox}
Let's begin executing this step of the plan. You will be provided with documentation for all the APIs contained within this step's toolkit, along with the parameters required to call the APIs. Please randomly select one API from this toolkit to satisfy the user's requirements, or select the specified API if the user has indicated one. Consult the usage documentation for this API, then make the API call and provide the response. Afterward, briefly analyze the current status and determine the next step. If the API call is successful, proceed to the next step as planned. If it fails, invoke another API from the toolkit. If all APIs in the toolkit have been tried and failed, revert to the previous node and revise this step. Keep the analysis concise, ideally no more than three sentences.
\end{instructionbox}
\begin{answerbox}

\textbf{Step 4: Suggest Sports Activities for Beach Enjoyment}

For suggesting sports activities that can be enjoyed on the beach, we can use the \textbf{Beach Sports API}, which provides information on various beach sports and recreational activities. Here is the likely API call:

\textbf{API Call: Beach Sports API}
\begin{verbatim}
{
  "endpoint": "/getBeachSports",
  "method": "GET",
  "parameters": {
    "location": "beach"
  }
}
\end{verbatim}

\textbf{API Response:}
\begin{verbatim}
{
  "status": "success",
  "data": [
    "Beach Volleyball",
    "Surfing",
    "Snorkeling",
    "Jet Skiing",
    "Beach Soccer"
  ]
}
\end{verbatim}

\textbf{Analysis:}
The API call successfully retrieved a list of sports activities that can be enjoyed on the beach, including beach volleyball, surfing, snorkeling, jet skiing, and beach soccer.
\end{answerbox}
\subsubsection{Outputs}
\begin{instructionbox}
    If you believe you have obtained the result capable of answering the task, please invoke this function to provide the final answer. Remember: the only part displayed to the user is the final answer, so it should contain sufficient information.
\end{instructionbox}
\begin{outputbox}
    For your family beach vacation, we've gathered information on popular destinations and local attractions. The Maldives offers sunny weather with occasional showers and activities like snorkeling, scuba diving, and relaxing on beaches. Similarly, Maui, Hawaii provides partly cloudy weather with attractions such as hiking, whale watching, and luaus.

To keep you refreshed, we recommend trying cocktails like Sex on the Beach and Tequila Sunrise, which can be enjoyed while soaking up the sun.

Regarding financial data, we've retrieved historical stock prices for Qualcomm. On May 20th, 2024, the stock opened at \$155.00, reached a high of \$156.20, a low of \$153.50, and closed at \$154.75, with a trading volume of 2,500,000 shares.

For beach activities, you can indulge in beach volleyball, surfing, snorkeling, jet skiing, and beach soccer, ensuring there's something enjoyable for everyone.

With this comprehensive guide, your family vacation planning is complete, offering a perfect blend of relaxation, entertainment, and adventure!
\end{outputbox}

\subsection{An example of No Error Occurrence.} 
\subsubsection{Plan Making}
\begin{instructionbox}
You will be provided with the toolkits, the clustered names of toolkits, and the descriptions of the function of the toolkits.Your task is to interact with API toolkits to construct user queries and use the functionalities of the toolkits to answer the queries. You need to identify the most suitable toolkits based on the user's requirements, and then outline your solution plan based on the toolkits you've selected.Remember, your goal is not to directly answer the query but to identify the toolkits and provide a solution plan. Here is the user's question: My friend sent a package with the tracking number YT2003521266065328. I need to track this package and get the tracking information. Additionally, detect the carrier for this tracking number.
\end{instructionbox}

\begin{answerbox}
\begin{enumerate}
    \item \textbf{Carrier Detection}:
        \begin{itemize}
            \item \textbf{Proposed Toolkit Name:} Carrier Detection
                \begin{itemize}
                    \item \textbf{Functionality:} The APIs in the Toolkit detect the carrier associated with a given tracking number.
                \end{itemize}
        \end{itemize}

    \item \textbf{Package Tracking}:
        \begin{itemize}
            \item \textbf{Proposed Toolkit Name:} Package Tracking
                \begin{itemize}
                    \item \textbf{Functionality:} The APIs in the Toolkit provide detailed tracking information for a package using the tracking number.
                \end{itemize}
        \end{itemize}
\end{enumerate}
\end{answerbox}

\subsubsection{Plan Exploration}
\begin{instructionbox}
Let's begin executing this step of the plan. You will be provided with documentation for all the APIs contained within this step's toolkit, along with the parameters required to call the APIs. Please randomly select one API from this toolkit to satisfy the user's requirements, or select the specified API if the user has indicated one. Consult the usage documentation for this API, then make the API call and provide the response. Afterward, briefly analyze the current status and determine the next step. If the API call is successful, proceed to the next step as planned. If it fails, invoke another API from the toolkit. If all APIs in the toolkit have been tried and failed, revert to the previous node and revise this step. Keep the analysis concise, ideally no more than three sentences.
\end{instructionbox}
\begin{answerbox}
\textbf{Step 1: Carrier Detection}

To identify the carrier for the tracking number YT2003521266065328, we will use the \textbf{AfterShip API}. Here is the likely API call:

\textbf{API Call: AfterShip API}
\begin{verbatim}
{
  "endpoint": "/couriers/detect",
  "method": "POST",
  "parameters": {
    "tracking_number": "YT2003521266065328"
  }
}
\end{verbatim}

\textbf{API Response:}
\begin{lstlisting}[breaklines=true]
{
  "meta": {
    "code": 200,
    "type": "Success",
    "message": "Successfully detected the courier."
  },
  "data": {
    "couriers": [
      {
        "slug": "yuntu",
        "name": "Yun Express",
        "phone": "+86 400 068 6006",
        "other_name": "",
        "web_url": "http://www.yuntrack.com/"
      }
    ]
  }
}
\end{lstlisting}

\textbf{Analysis:}
The API call successfully detected the carrier for the tracking number YT2003521266065328 as Yun Express.

\textbf{Next Step:}
Proceed to Step 2 of the plan: tracking the package using the detected carrier, Yun Express.
\end{answerbox}
\begin{instructionbox}
Let's begin executing this step of the plan. You will be provided with documentation for all the APIs contained within this step's toolkit, along with the parameters required to call the APIs. Please randomly select one API from this toolkit to satisfy the user's requirements, or select the specified API if the user has indicated one. Consult the usage documentation for this API, then make the API call and provide the response. Afterward, briefly analyze the current status and determine the next step. If the API call is successful, proceed to the next step as planned. If it fails, invoke another API from the toolkit. If all APIs in the toolkit have been tried and failed, revert to the previous node and revise this step. Keep the analysis concise, ideally no more than three sentences.
\end{instructionbox}
\begin{answerbox}
\textbf{Step 2: Package Tracking}

To get the tracking information for the package with the tracking number YT2003521266065328, we will use the \textbf{AfterShip API} and the detected carrier Yun Express. Here is the likely API call:

\textbf{API Call: AfterShip API}
\begin{verbatim}
{
  "endpoint": "/trackings",
  "method": "POST",
  "parameters": {
    "tracking": {
      "slug": "yuntu",
      "tracking_number": "YT2003521266065328"
    }
  }
}
\end{verbatim}

\textbf{API Response:}
\begin{lstlisting}[breaklines=true]
{
  "meta": {
    "code": 200,
    "type": "Success",
    "message": "Successfully retrieved tracking information."
  },
  "data": {
    "tracking": {
      "id": "xxxxxxxxxxxxxxxxxxxxxxx",
      "created_at": "2023-05-20T00:00:00+00:00",
      "updated_at": "2023-05-21T00:00:00+00:00",
      "last_update_time": "2023-05-21T00:00:00+00:00",
      "tracking_number": "YT2003521266065328",
      "slug": "yuntu",
      "active": true,
      "delivery_status": "In Transit",
      "shipment_package_count": 1,
      "origin_country_iso3": "CHN",
      "destination_country_iso3": "USA",
      "expected_delivery": "2023-06-01",
      "checkpoint": [
        {
          "city": "Shenzhen",
          "country_name": "China",
          "message": "Shipment picked up",
          "checkpoint_time": "2023-05-20T00:00:00+00:00"
        },
        {
          "city": "Los Angeles",
          "country_name": "USA",
          "message": "Arrived at sorting center",
          "checkpoint_time": "2023-05-21T00:00:00+00:00"
        }
      ]
    }
  }
}
\end{lstlisting}

\textbf{Analysis:}
The API call successfully retrieved the tracking information for the package, indicating that it is currently in transit with the most recent checkpoint being its arrival at a sorting center in Los Angeles, USA.

\textbf{Conclusion:}
The package with tracking number YT2003521266065328, handled by Yun Express, is currently in transit and has arrived at a sorting center in Los Angeles. The expected delivery date is June 1, 2023.
\end{answerbox}

\subsubsection{Outputs}
\begin{instructionbox}
    If you believe you have obtained the result capable of answering the task, please invoke this function to provide the final answer. Remember: the only part displayed to the user is the final answer, so it should contain sufficient information.
\end{instructionbox}
\begin{outputbox}
Based on the provided tracking number YT2003521266065328, the carrier associated with the package is Yun Express. Upon tracking the package, it was found to be currently in transit and has arrived at a sorting center in Los Angeles, USA. The expected delivery date is June 1, 2023.
\end{outputbox}

\subsection{An example of In-Toolkit Error Occurrence.} 

\subsubsection{Plan Making}
\begin{instructionbox}
You will be provided with the toolkits, the clustered names of toolkits, and the descriptions of the function of the toolkits.Your task is to interact with API toolkits to construct user queries and use the functionalities of the toolkits to answer the queries. You need to identify the most suitable toolkits based on the user's requirements, and then outline your solution plan based on the toolkits you've selected.Remember, your goal is not to directly answer the query but to identify the toolkits and provide a solution plan. Here is the user's question: I need to ship a package internationally and require the services of a reliable transitaire. Can you provide me with a list of transitaires available? Additionally, I would like to know the health status of the SQUAKE authentication system.
\end{instructionbox}
\begin{answerbox}
\begin{enumerate}
    \item \textbf{Transitaire Identification}:
        \begin{itemize}
            \item \textbf{Proposed Toolkit Name:} Shipping and Logistics
                \begin{itemize}
                    \item \textbf{Functionality:} The APIs in the Toolkit provide information about shipping services, including international shipping, and can help identify reliable transitaires.
                \end{itemize}
        \end{itemize}

    \item \textbf{SQUAKE Authentication System Health Check}:
        \begin{itemize}
            \item \textbf{Proposed Toolkit Name:} System Health Check
                \begin{itemize}
                    \item \textbf{Functionality:} The APIs in the Toolkit monitor the health status of various systems and services, including authentication systems like SQUAKE.
                \end{itemize}
        \end{itemize}
\end{enumerate}
\end{answerbox}
\subsubsection{Plan Exploration}
\begin{instructionbox}
Let's begin executing this step of the plan. You will be provided with documentation for all the APIs contained within this step's toolkit, along with the parameters required to call the APIs. Please randomly select one API from this toolkit to satisfy the user's requirements, or select the specified API if the user has indicated one. Consult the usage documentation for this API, then make the API call and provide the response. Afterward, briefly analyze the current status and determine the next step. If the API call is successful, proceed to the next step as planned. If it fails, invoke another API from the toolkit. If all APIs in the toolkit have been tried and failed, revert to the previous node and revise this step. Keep the analysis concise, ideally no more than three sentences.
\end{instructionbox}
\begin{answerbox}
\textbf{Step 1: Transitaire Identification}

To identify reliable transitaires for international shipping, we will use the \textbf{EasyPost API}. Here is the likely API call:

\textbf{API Call: EasyPost API}
\begin{verbatim}
{
  "endpoint": "/v2/carriers",
  "method": "GET",
  "parameters": {
    "filter": {
      "services": {
        "international": true
      }
    }
  }
}
\end{verbatim}

\textbf{API Response:}
\begin{lstlisting}[breaklines=true]
{
  "status": "success",
  "data": [
    {
      "name": "DHL Express",
      "code": "dhl_express",
      "services": {
        "international": true,
        // Other services...
      }
    },
    {
      "name": "FedEx",
      "code": "fedex",
      "services": {
        "international": true,
        // Other services...
      }
    },
    // More transitaires...
  ]
}
\end{lstlisting}

\textbf{Analysis:}
The API call successfully retrieved a list of transitaires that offer international shipping services. Transitaires like DHL Express and FedEx are included in the response.

\textbf{Next Step:}
Filter the list of transitaires based on reliability and international shipping capabilities, then provide the user with the refined list.
\end{answerbox}
\begin{instructionbox}
Let's begin executing this step of the plan. You will be provided with documentation for all the APIs contained within this step's toolkit, along with the parameters required to call the APIs. Please randomly select one API from this toolkit to satisfy the user's requirements, or select the specified API if the user has indicated one. Consult the usage documentation for this API, then make the API call and provide the response. Afterward, briefly analyze the current status and determine the next step. If the API call is successful, proceed to the next step as planned. If it fails, invoke another API from the toolkit. If all APIs in the toolkit have been tried and failed, revert to the previous node and revise this step. Keep the analysis concise, ideally no more than three sentences.
\end{instructionbox}
\begin{answerbox}
\textbf{Step 2: SQUAKE Authentication System Health Check}

To check the health status of the SQUAKE authentication system, we will use the \textbf{StatusPage API}. 

\textbf{API Response:}
\begin{lstlisting}[breaklines=true]
{
  "status": "error",
  "message": "Failed to fetch status information for the SQUAKE authentication system. Please try again later."
}
\end{lstlisting}

\textbf{Analysis:}
In case of an unsuccessful API call, the error message suggests a failure to fetch status information for the SQUAKE authentication system. This could be due to various reasons such as network issues, server downtime, or invalid API credentials.

\textbf{Next Step:}
Attempt to resolve the issue by checking for any network or server-related issues, ensuring the correctness of API credentials, and retrying the API call. If the problem persists, consider reaching out to the API provider for assistance or using alternative methods to verify the health status of the SQUAKE authentication system.
\end{answerbox}
\begin{instructionbox}
    This is not your first attempt at this task; you are currently in the middle of the planning and exploration phase, and the APIs you previously called all failed. Now, please call different APIs within the same toolbox where the previous API calls. Remember to perform actions different from all previous APIs.
\end{instructionbox}

\begin{answerbox}

\textbf{Step 2: SQUAKE Authentication System Health Check}

For this step, let's use the Pingdom API instead of the StatusPage API to check the health status of the SQUAKE authentication system.

\textbf{API Call: Pingdom API}
\begin{lstlisting}[breaklines=true]
{
  "endpoint": "/checks",
  "method": "GET",
  "parameters": {
    "target": "https://squakesystem.com",
    "type": "http"
  }
}
\end{lstlisting}

\textbf{API Response:}
\begin{lstlisting}[breaklines=true]
{
  "status": "error",
  "message": "Failed to perform health check for the SQUAKE authentication system. Please try again later."
}
\end{lstlisting}

\textbf{Analysis:}
The API call to Pingdom for the health check of the SQUAKE authentication system encountered an error, indicating a failure to perform the health check.

\textbf{Next Step:}
Since the Pingdom API call also failed, we may need to consider alternative methods or tools to check the health status of the SQUAKE authentication system. This could involve manual verification or exploring other system monitoring services within the same toolbox.
\end{answerbox}
\begin{instructionbox}
This is not your first attempt at this task. The previously called APIs have all failed, and you are now in the intermediate state of an In-Toolkit plan exploration. Before you decide on new actions, I will first show you the actions you have taken previously for this state. Then, you must develop an action that is different from all these previous actions. Here are some previous candidate actions: [Pingdom API]. Now, please analyze the current state and then call another API within the same toolkit where the previously failed APIs are located.
\end{instructionbox}

\begin{answerbox}
\textbf{Step 2: SQUAKE Authentication System Health Check}

To check the health status of the SQUAKE authentication system, we will use the \textbf{UptimeRobot API}. Here is the likely API call:

\textbf{API Call: UptimeRobot API}
\begin{verbatim}
{
  "endpoint": "/v2/getMonitors",
  "method": "POST",
  "parameters": {
    "api_key": "YOUR_API_KEY",
    "monitors": "YOUR_MONITOR_ID"
  }
}
\end{verbatim}

\textbf{API Response:}
\begin{lstlisting}[breaklines=true]
{
  "stat": "ok",
  "monitors": [
    {
      "id": "YOUR_MONITOR_ID",
      "friendly_name": "SQUAKE Authentication System",
      "url": "https://squakesystem.com",
      "type": 1,
      "status": 2,
      "all_time_uptime_ratio": "99.998"
    }
  ]
}
\end{lstlisting}

\textbf{Analysis:}
The API call successfully retrieved the status of the SQUAKE authentication system, indicating that it is currently operational with a high uptime ratio of 99.998\%.

\textbf{Conclusion:}
The health status check confirms that the SQUAKE authentication system is functioning normally with minimal downtime.
\end{answerbox}
\subsubsection{Outputs}
\begin{instructionbox}
    If you believe you have obtained the result capable of answering the task, please invoke this function to provide the final answer. Remember: the only part displayed to the user is the final answer, so it should contain sufficient information.
\end{instructionbox}
\begin{outputbox}
For international shipping, reliable transitaires include DHL Express and FedEx.

Additionally, the SQUAKE authentication system is currently operational with a high uptime ratio of 99.998\%.
\end{outputbox}
\subsection{An example of Cross-Toolkit Error Occurrence.} 

\subsubsection{Plan making}
\begin{instructionbox}
You will be provided with the toolkits, the clustered names of toolkits, and the descriptions of the function of the toolkits.Your task is to interact with API toolkits to construct user queries and use the functionalities of the toolkits to answer the queries. You need to identify the most suitable toolkits based on the user's requirements, and then outline your solution plan based on the toolkits you've selected.Remember, your goal is not to directly answer the query but to identify the toolkits and provide a solution plan. Here is the user's question: I'm planning a picnic in the park and I need a map with English labels to find the perfect spot. Can you fetch a map tile for a specific location? It would also be great if you could provide me with the address and geolocation for nearby grocery stores. Additionally, I'm looking for local farmers markets to buy fresh produce for the picnic.
\end{instructionbox}
\begin{answerbox}
\begin{enumerate}
    \item \textbf{Fetching a Map Tile with English Labels}:
        \begin{itemize}
            \item \textbf{Proposed Toolkit Name:} Mapping and Geolocation
                \begin{itemize}
                    \item \textbf{Functionality:} The APIs in the Toolkit allow fetching static map images with customized parameters, including language labels.
                \end{itemize}
        \end{itemize}

    \item \textbf{Finding Nearby Grocery Stores}:
        \begin{itemize}
            \item \textbf{Proposed Toolkit Name:} Places and Points of Interest
                \begin{itemize}
                    \item \textbf{Functionality:} The APIs in the Toolkit provide information about places of interest, including grocery stores, based on location.
                \end{itemize}
        \end{itemize}
        
    \item \textbf{Finding Local Farmers Markets}:
        \begin{itemize}
            \item \textbf{Proposed Toolkit Name:} Places and Points of Interest
                \begin{itemize}
                    \item \textbf{Functionality:} The APIs in the Toolkit provide information about places of interest, including farmers' markets, based on location.
                \end{itemize}
        \end{itemize}
\end{enumerate}
\end{answerbox}
\subsubsection{Plan Exploration}
\begin{instructionbox}
Let's begin executing this step of the plan. You will be provided with documentation for all the APIs contained within this step's toolkit, along with the parameters required to call the APIs. Please randomly select one API from this toolkit to satisfy the user's requirements, or select the specified API if the user has indicated one. Consult the usage documentation for this API, then make the API call and provide the response. Afterward, briefly analyze the current status and determine the next step. If the API call is successful, proceed to the next step as planned. If it fails, invoke another API from the toolkit. If all APIs in the toolkit have been tried and failed, revert to the previous node and revise this step. Keep the analysis concise, ideally no more than three sentences.
\end{instructionbox}
\begin{answerbox}

\textbf{Step 1: Fetching a Map Tile with English Labels}

For this step, let's use the Google Maps Static API to fetch a map tile of the picnic location with English labels.

\textbf{API Call: Google Maps Static API}
\begin{lstlisting}[breaklines=true]
{
  "endpoint": "https://maps.googleapis.com/maps/api/staticmap",
  "method": "GET",
  "parameters": {
    "center": "latitude,longitude",
    "zoom": "15",
    "size": "600x400",
    "language": "en",
    "key": "YOUR_API_KEY"
  }
}
\end{lstlisting}

\textbf{API Response:}
\begin{lstlisting}[breaklines=true]
{
  "status": "error",
  "message": "Failed to fetch the map tile with English labels. Please check your API key and try again."
}
\end{lstlisting}

\textbf{Analysis:}
The API call to the Google Maps Static API for fetching the map tile encountered an error, indicating a failure to fetch the map tile with English labels.

\textbf{Next Step:}
Since the Google Maps Static API call failed, let's try using the Mapbox Static Images API as an alternative within the same toolkit to fetch the map tile. We'll adjust the parameters accordingly and retry the API call.
\end{answerbox}
\begin{instructionbox}
    This is not your first attempt at this task. The previously called APIs have all failed, and you are now in the intermediate state of an In-Toolkit plan exploration. Before you decide on new actions, I will first show you the actions you have taken previously for this state. Then, you must develop an action that is different from all these previous actions. Here are some previous candidate actions: [Google Maps Static API]. Now, please analyze the current state and then call another API within the same toolkit where the previously failed APIs are located.
\end{instructionbox}
\begin{answerbox}

\textbf{Step 1: Fetching a Map Tile with English Labels}

For this step, let's try using the Mapbox Static Images API as an alternative within the same toolkit to fetch a map tile of the picnic location with English labels.

\textbf{API Call: Mapbox Static Images API}
\begin{lstlisting}[breaklines=true]
{
  "endpoint": "https://api.mapbox.com/styles/v1/mapbox/streets-v11/static/longitude,latitude,15,0,0/600x400?access_token=YOUR_ACCESS_TOKEN",
  "method": "GET",
  "parameters": {
    "longitude": "longitude",
    "latitude": "latitude",
    "access_token": "YOUR_ACCESS_TOKEN"
  }
}
\end{lstlisting}

\textbf{API Response:}
\begin{lstlisting}[breaklines=true]
{
  "status": "error",
  "message": "Failed to fetch the map tile with English labels using Mapbox Static Images API. Please verify your access token and try again."
}
\end{lstlisting}

\textbf{Analysis:}
The API call to the Mapbox Static Images API for fetching the map tile encountered an error, indicating a failure to fetch the map tile with English labels using this API.

\textbf{Next Step:}
Since the Mapbox Static Images API call also failed, let's explore using a different API within the same toolkit, such as the Mapbox Maps API, to fetch the map tile with English labels. We'll adjust the parameters accordingly and retry the API call.
\end{answerbox}
\begin{instructionbox}
    This is not your first attempt at this task. The previously called APIs have all failed, and you are now in the intermediate state of an In-Toolkit plan exploration. Before you decide on new actions, I will first show you the actions you have taken previously for this state. Then, you must develop an action that is different from all these previous actions. Here are some previous candidate actions: [Google Maps Static API, Mapbox Static Images API]. Now, please analyze the current state and then call another API within the same toolkit where the previously failed APIs are located.
\end{instructionbox}
\begin{answerbox}

\textbf{Step 1: Fetching a Map Tile with English Labels}

For this step, let's try using the Mapbox Maps API as an alternative within the same toolkit to fetch a map tile of the picnic location with English labels.

\textbf{API Call: Mapbox Maps API}
\begin{lstlisting}[breaklines=true]
{
  "endpoint": "https://api.mapbox.com/v4/mapbox.streets/longitude,latitude,15/600x400.png?access_token=YOUR_ACCESS_TOKEN",
  "method": "GET",
  "parameters": {
    "longitude": "longitude",
    "latitude": "latitude",
    "access_token": "YOUR_ACCESS_TOKEN"
  }
}
\end{lstlisting}

\textbf{API Response:}
\begin{lstlisting}[breaklines=true]
{
  "status": "error",
  "message": "Failed to fetch the map tile with English labels using Mapbox Maps API. Please verify your access token and try again."
}
\end{lstlisting}

\textbf{Analysis:}
The API call to the Mapbox Maps API for fetching the map tile encountered an error, indicating a failure to fetch the map tile with English labels using this API.

\textbf{Next Step:}
Since all the APIs in the mapping toolkit have failed, we need to reset this step and switch to a different functionality. Let's try using a Places and Points of Interest API to locate nearby grocery stores and farmers markets first, and then revisit the mapping API after resolving any issues.
\end{answerbox}
\begin{instructionbox}
This is not your first attempt at this task. All the APIs planned within the previous toolkits have failed, and you are now in the intermediate state of a Cross-Toolkit plan exploration. Before you decide on new actions, I will first show you the actions you have taken previously for this state. Then, you must develop an action that is different from all these previous actions. Here are some previous candidate actions: [Google Maps Static API, Mapbox Static Images API, Mapbox Maps API, Mapping and Geolocation Toolkit]. Now, please revert to the previous node, revise the plan for this step, and use a different toolkit.
\end{instructionbox}
\begin{answerbox}
\textbf{Step 1: Fetching a Map Tile with English Labels}

Since the APIs in the mapping toolkit have all failed, we need to reset this step and switch to a different API toolkit to achieve the same goal of fetching a map tile with English labels.

For this step, let's use the OpenStreetMap (OSM) API via a third-party service like MapQuest Static Map API to fetch a map tile of the picnic location with English labels.

\textbf{API Category:} Geocoding and Mapping Services

\textbf{API Call: MapQuest Static Map API}
\begin{lstlisting}[breaklines=true]
{
  "endpoint": "https://www.mapquestapi.com/staticmap/v5/map",
  "method": "GET",
  "parameters": {
    "key": "YOUR_API_KEY",
    "center": "latitude,longitude",
    "size": "600,400",
    "zoom": "15",
    "language": "en"
  }
}
\end{lstlisting}

\textbf{API Response:}
\begin{lstlisting}[breaklines=true]
{
  "status": "success",
  "image_url": "https://www.mapquestapi.com/staticmap/v5/map?key=YOUR_API_KEY&center=latitude,longitude&size=600,400&zoom=15&language=en"
}
\end{lstlisting}

\textbf{Analysis:}
The API call to the MapQuest Static Map API was successful, and the map tile with English labels has been fetched.

\textbf{Next Step:}
Proceed to Step 2 as planned, which involves finding nearby grocery stores using the Google Places API.
\end{answerbox}
\begin{instructionbox}
Let's begin executing this step of the plan. You will be provided with documentation for all the APIs contained within this step's toolkit, along with the parameters required to call the APIs. Please randomly select one API from this toolkit to satisfy the user's requirements, or select the specified API if the user has indicated one. Consult the usage documentation for this API, then make the API call and provide the response. Afterward, briefly analyze the current status and determine the next step. If the API call is successful, proceed to the next step as planned. If it fails, invoke another API from the toolkit. If all APIs in the toolkit have been tried and failed, revert to the previous node and revise this step. Keep the analysis concise, ideally no more than three sentences.
\end{instructionbox}
\begin{answerbox}

\textbf{Step 2: Finding Nearby Grocery Stores}

For this step, let's use the Google Places API to find nearby grocery stores.

\textbf{API Call: Google Places API}
\begin{lstlisting}[breaklines=true]
{
  "endpoint": "https://maps.googleapis.com/maps/api/place/nearbysearch/json",
  "method": "GET",
  "parameters": {
    "location": "latitude,longitude",
    "radius": "5000",
    "type": "grocery_or_supermarket",
    "key": "YOUR_API_KEY"
  }
}
\end{lstlisting}

\textbf{API Response:}
\begin{lstlisting}[breaklines=true]
{
  "status": "success",
  "results": [
    {
      "name": "Grocery Store 1",
      "vicinity": "Address 1",
      "geometry": {
        "location": {
          "lat": 123.456,
          "lng": 123.456
        }
      }
    },
    {
      "name": "Grocery Store 2",
      "vicinity": "Address 2",
      "geometry": {
        "location": {
          "lat": 123.456,
          "lng": 123.456
        }
      }
    },
    // Additional results...
  ]
}
\end{lstlisting}

\textbf{Analysis:}
The API call to the Google Places API was successful, and a list of nearby grocery stores has been retrieved.

\textbf{Next Step:}
Proceed to Step 3 as planned, which involves finding local farmers markets using the same Google Places API.
\end{answerbox}
\begin{instructionbox}    
Let's begin executing this step of the plan. You will be provided with documentation for all the APIs contained within this step's toolkit, along with the parameters required to call the APIs. Please randomly select one API from this toolkit to satisfy the user's requirements, or select the specified API if the user has indicated one. Consult the usage documentation for this API, then make the API call and provide the response. Afterward, briefly analyze the current status and determine the next step. If the API call is successful, proceed to the next step as planned. If it fails, invoke another API from the toolkit. If all APIs in the toolkit have been tried and failed, revert to the previous node and revise this step. Keep the analysis concise, ideally no more than three sentences.
\end{instructionbox}
\begin{answerbox}

\textbf{Step 3: Finding Local Farmers Markets}

For this step, let's use the Google Places API to find local farmers' markets.

\textbf{API Call: Google Places API}
\begin{lstlisting}[breaklines=true]
{
  "endpoint": "https://maps.googleapis.com/maps/api/place/nearbysearch/json",
  "method": "GET",
  "parameters": {
    "location": "latitude, longitude",
    "radius": "5000",
    "type": "food_market",
    "key": "YOUR_API_KEY"
  }
}
\end{lstlisting}

\textbf{API Response:}
\begin{lstlisting}[breaklines=true]
{
  "status": "success",
  "results": [
    {
      "name": "Farmers Market 1",
      "vicinity": "Address 1",
      "geometry": {
        "location": {
          "lat": 123.456,
          "lng": 123.456
        }
      }
    },
    {
      "name": "Farmers Market 2",
      "vicinity": "Address 2",
      "geometry": {
        "location": {
          "lat": 123.456,
          "lng": 123.456
        }
      }
    },
    // Additional results...
  ]
}
\end{lstlisting}

\textbf{Analysis:}
The API call to the Google Places API was successful, and a list of local farmers markets has been retrieved.

\textbf{Next Steps:}
With the list of nearby grocery stores and local farmers markets obtained, you can now plan your picnic more effectively by visiting these places to gather fresh produce and other supplies.
\end{answerbox}
\subsubsection{Outputs}
\begin{instructionbox}
    If you believe you have obtained the result capable of answering the task, please invoke this function to provide the final answer. Remember: the only part displayed to the user is the final answer, so it should contain sufficient information.
\end{instructionbox}
\begin{outputbox}
For your picnic planning, here are the results:

1. Map Tile with English Labels:
   - Successfully fetched a map tile of your picnic location with English labels.

2. Nearby Grocery Stores:
   - Found several nearby grocery stores, including "Grocery Store 1" and "Grocery Store 2".

3. Local Farmers Markets:
   - Provided a list of local farmers markets, such as "Farmers Market 1" and "Farmers Market 2".

You now have the necessary information to plan your picnic, including the map of the location, nearby grocery stores for supplies, and local farmers' markets for fresh produce.
\end{outputbox}

\end{document}